\documentclass[journal]{IEEEtran}

\usepackage{amsmath,amssymb,amsfonts}
\usepackage[inline]{enumitem}
\usepackage{algorithmic}
\usepackage{graphicx}
\usepackage{textcomp}
\usepackage{adjustbox}
\usepackage{booktabs}
\usepackage{multirow}
\usepackage{hyperref}
\hypersetup{
colorlinks=true,linkcolor=red,citecolor=rblue,urlcolor=black}

\usepackage{tabulary}
\usepackage[table]{xcolor}

\definecolor{gray}{rgb}{0.3,0.3,0.3}
\definecolor{yellow}{rgb}{0.94,0.90,0.25}
\definecolor{blue}{rgb}{0,0.5,1}
\definecolor{green}{rgb}{0.2,1,0.2}
\definecolor{rblue}{rgb}{0,0,1}

\newcommand{\green}[1]{\textcolor[RGB]{96,177,87}{#1}}
\newcommand{\fn}[1]{\footnotesize{#1}}
\newcommand{\gbf}[1]{\green{\bf{\fn{(#1)}}}}

\usepackage{subcaption}

\usepackage[capitalise]{cleveref}

\begin{document}
\title{Exploring Event-driven Dynamic Context for Accident Scene Segmentation}
\author{Jiaming Zhang, Kailun Yang, and Rainer Stiefelhagen

\thanks{This work was supported in part by the AccessibleMaps Project by the Federal Ministry of Labor and Social Affairs (BMAS) under Grant 01KM151112, in part by the University of Excellence through the “KIT Future Fields” project, and in part by Hangzhou SurImage Company Ltd.  (\textit{Corresponding author: Kailun Yang).}}
\thanks{The authors are with the Institute for Anthropomatics and Robotics, Karlsruhe Institute of Technology, 76131 Karlsruhe, Germany (e-mail: jiaming.zhang@kit.edu; kailun.yang@kit.edu; rainer.stiefelhagen@kit.edu).}
\thanks{{Code and dataset will be made publicly available at:~\url{https://github.com/jamycheung/ISSAFE}}}
}
\maketitle

\begin{abstract}
The robustness of semantic segmentation on edge cases of traffic scene is a vital factor for the safety of intelligent transportation. However, most of the critical scenes of traffic accidents are extremely dynamic and previously unseen, which seriously harm the performance of semantic segmentation methods. In addition, the delay of the traditional camera during high-speed driving will further reduce the contextual information in the time dimension. Therefore, we propose to extract dynamic context from event-based data with a higher temporal resolution to enhance static RGB images, even for those from traffic accidents with motion blur, collisions, deformations, overturns, \textit{etc.} Moreover, in order to evaluate the segmentation performance in traffic accidents, we provide a pixel-wise annotated accident dataset, namely DADA-seg, which contains a variety of critical scenarios from traffic accidents. Our experiments indicate that event-based data can provide complementary information to stabilize semantic segmentation under adverse conditions by preserving fine-grained motion of fast-moving foreground (crash objects) in accidents. Our approach achieves $+8.2\%$ performance gain on the proposed accident dataset, exceeding more than $20$ state-of-the-art semantic segmentation methods. The proposal has been demonstrated to be consistently effective for models learned on multiple source databases including Cityscapes, KITTI-360, BDD, and ApolloScape. 

\end{abstract}

\begin{IEEEkeywords}
Semantic segmentation, scene understanding, accident scenarios, event-based vision, intelligent vehicles.
\end{IEEEkeywords}

\IEEEpeerreviewmaketitle

\section{Introduction}
\IEEEPARstart{T}{he} road safety of Intelligent Vehicles (IV) is being concurrently questioned and researched intensely along with the rapid development of self-driving systems.
The general pipeline of self-driving systems can be divided into four sections including (a) the sensing module to observe the environment~\cite{sun2019multimodal}; (b) the perception module that uses captured data from the sensors to offer high-level information like recognition results about the surrounding scenes~\cite{muresan2020stabilization}; (c) the path planning module which plans the motion of the vehicle based on the sensing and perception results~\cite{kim2017intervention}; and (d) the control module that commands the vehicle to make certain actions~\cite{arikan2018control}.
In recent decades, Intelligent Transportation Systems (ITS) have been growing at a rapid pace based on the development of intelligent perception technology, which serves as a vital and basic component of self-driving systems and the quality of scene understanding will directly affect the decision-making process. 
In particular, computer vision models learned on real-world data not only enhance the performance of the perception component but also help to comprehensively understand the surrounding environments~\cite{cordts2016cityscapes}. Most recently, to improve road safety, some works applied vision-based sensing technologies to perform detection~\cite{le2020attention_accidentdetection}\cite{pashaei2020accident_classify}\cite{you2020traffic_accident_benchmark} and prediction~\cite{yuan2018accident_prediction} of traffic accidents, as well as estimation of driver attention~\cite{FangDADA}.

\begin{figure}
    \centering
	\includegraphics[width=0.99\columnwidth]{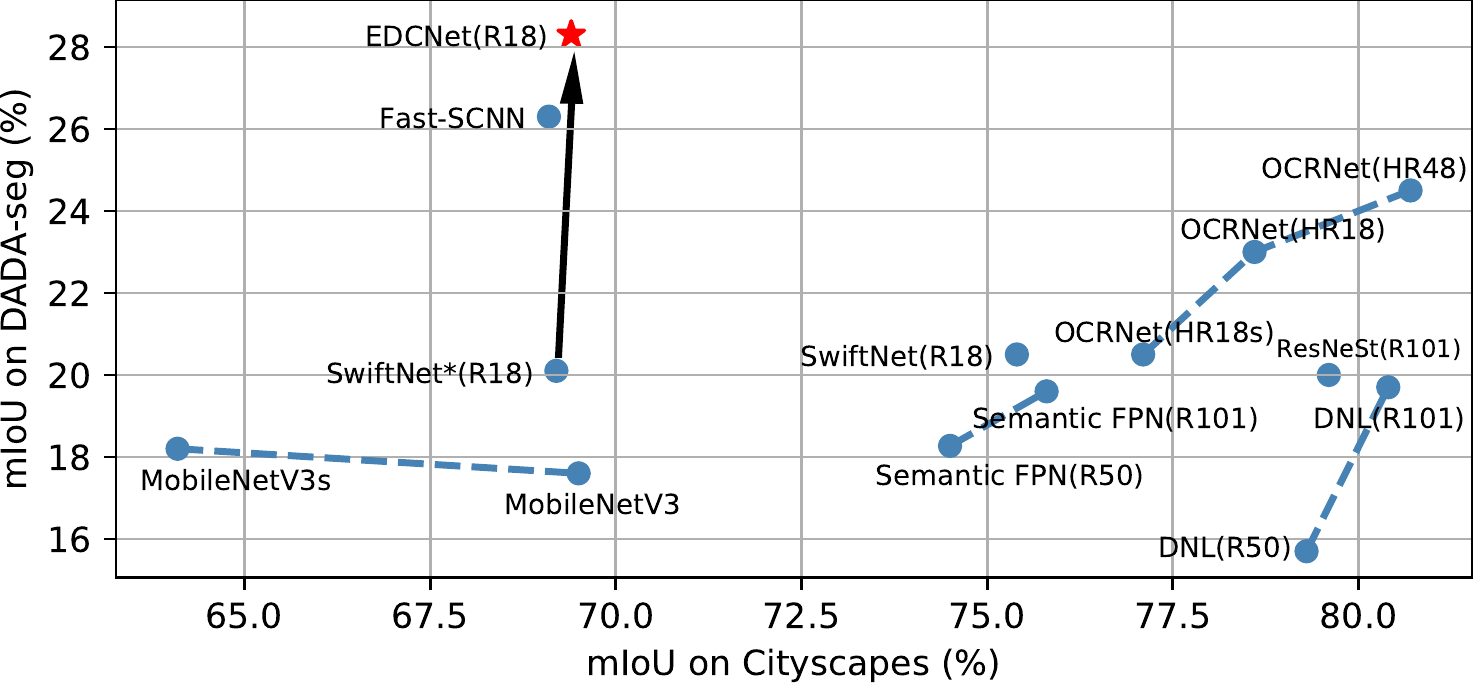}
	\vskip-1ex
	\caption{\small Segmentation accuracy in mIoU(\%) across Cityscapes validation and DADA-seg testing datasets. The red star denotes the accuracy of our proposed EDCNet. It achieves $+8.2\%$ gain compared to SwiftNet* which is trained from scratch under the same setting. The mark $s$ represents the corresponding small version of the network.}
    \label{fig:miou_gap}
\vskip -1ex
\end{figure}
\begin{figure}
    \centering
    \includegraphics[width=0.99\columnwidth]{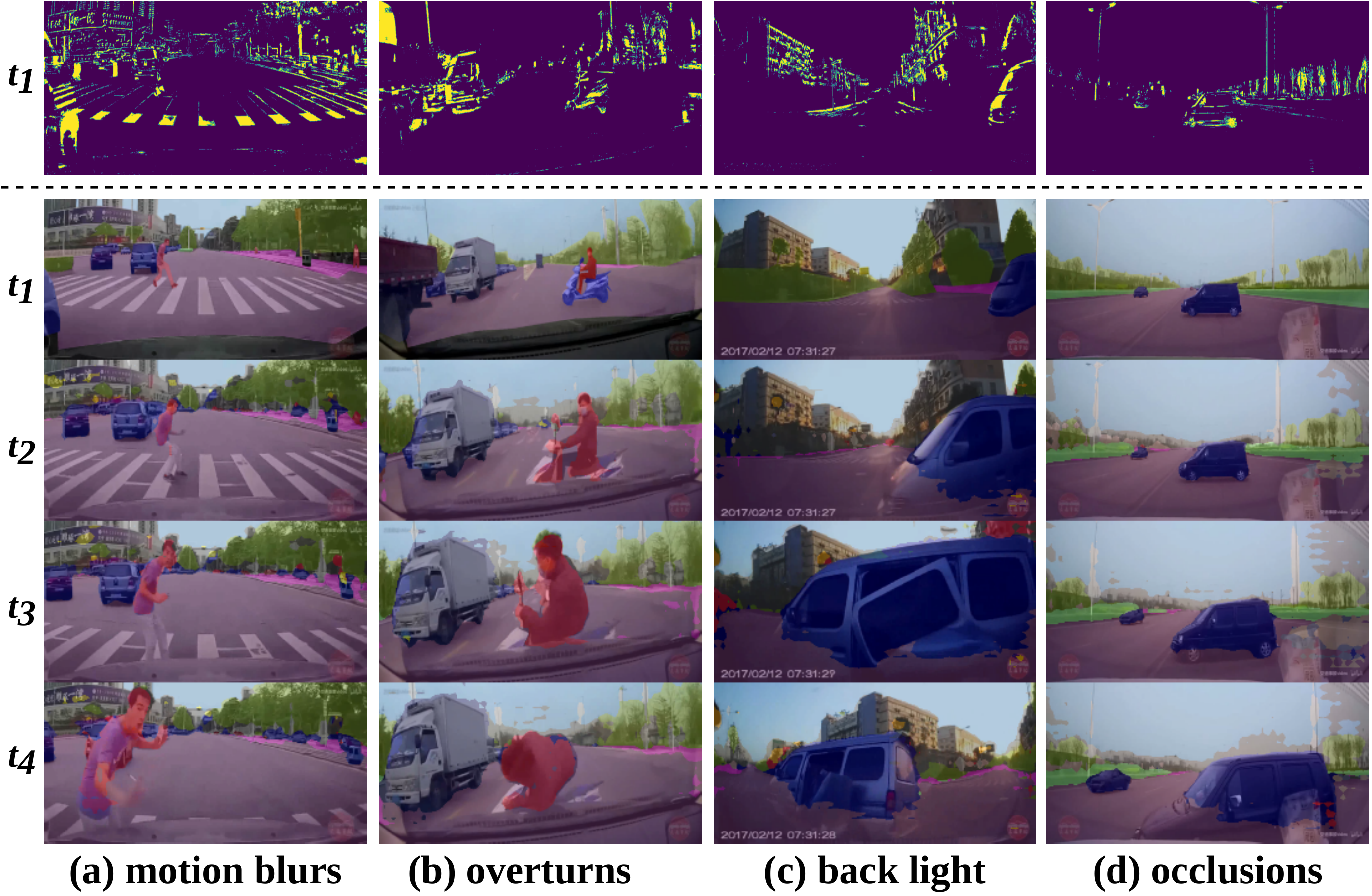}
	\vskip -1ex
    \caption{\small Accident sequences from the proposed \emph{DADA-seg} dataset include diverse hazards (\textit{e.g.}, motion blur, overturns, back light, object occlusions). From top to bottom: event data of the $t_1$ frame, and four frames before and during an accident, where the $t_{1}$ frame is the ground-truth segmentation for quantitative evaluation, and the others are predictions of our EDCNet model. Zoom in for a better view.}
    \label{fig:DADA_SegExample}
\vskip -3ex
\end{figure}

In contrast, image semantic segmentation is generally regarded as an ideal solution for parsing complicated street scenes by providing a unified and dense image classification, containing object categories, shapes, and locations~\cite{feng2020deep}. In recent years, many state-of-the-art models~\cite{zhao2017PSPNet}\cite{chen2018deeplabv3plus}\cite{yuan2019OCRNet}\cite{fu2019dual}\cite{yin2020dnl} have achieved impressive successes 
on major segmentation benchmarks~\cite{cordts2016cityscapes}\cite{xie2016semantic}\cite{yu2020bdd100k}\cite{wang2019apolloscape}.
However, the research of applying semantic segmentation methods on traffic accident images is relatively rare.
In addition, a direct transfer of trained models from common street-scene images to those containing traffic accidents will mostly result in a significant performance drop (Fig.~\ref{fig:miou_gap}). 
Thus, this work concentrates on exploring the robustness of semantic segmentation models in processing edge cases during driving, in particular, traffic accident images.

For this purpose, we create an alternative benchmark \emph{DADA-seg} {based on the large-scale DADA-2000 database~\cite{FangDADA}} for a new task, \emph{i.e. accident scene segmentation}. To facilitate this research as well as to supplement existing benchmarks~\cite{cordts2016cityscapes}\cite{xie2016semantic}\cite{yu2020bdd100k}\cite{wang2019apolloscape}\cite{zendel2018wilddash}, we propose an associated dataset which is collected from real-world traffic accidents, involving highly dynamic scenarios and extremely adverse factors. Some cases are shown in \cref{fig:DADA_SegExample}, covering diverse situations: motion blur while the pedestrian is dashing across the road, overturning of the motorcyclist during the collision, back-lighting at the intersection, and the occlusions by windshield reflection. As far as known to us, these factors are still challenging for most existing segmentation algorithms and harmful for their performance. Note that the main objective of creating this benchmark is to provide a set of edge cases (critical and accidental) for testing the robustness of models before deployment. {Though the direct prediction of traffic accidents is excluded in our scope, the robustness of the semantic segmentation model is a vital factor in IV systems.} 

Fusing sensors in an IV system can obtain various data modalities at the sensing stage~\cite{sun2019multimodal}. While LiDAR finely scans objects with laser light pulses in a certain range, the RADAR sensor uses radio waves to scan objects in a longer range. However, the LiDAR scans are inaccurate in severe weather that often leads to traffic accidents. RADAR is not affected by weather conditions but has a lower resolution and precision of measurement compared to LiDAR, especially for the perception of small objects. These sensors acquire visual data by discrete time stamps similar to traditional cameras. Therefore, visual information is quantified at a predefined frame rate, which results in the loss of information between adjacent frames~\cite{chen2020event}. In addition, the price and size of LiDAR also hinder its wide deployment to IV systems. 
Unlike traditional frame-based sensors, event cameras are bio-inspired novel sensors, such as the Dynamic Vision Sensor (DVS)~\cite{patrick2008DVS_sensor}, that encode \emph{changes} of intensity at each pixel asynchronously. Thanks to the characteristics of (a) higher dynamic range ($>120 dB$), (b) high time resolution ($1 MHz$ clock or $\mu s$ timestamp), (c) lower power consumption, and (d) robustness against motion blur~\cite{gallego2019event_survey}, event-based neuromorphic vision sensors constitute a novel paradigm for perceiving driving scenes.
Apart from the higher dynamic range for enhancing overexposed or underexposed scenes, event cameras with higher time resolutions are more sensitive to capture motion information during dynamic driving, especially for fast-moving objects (foreground classes) in accident scenarios, where classic cameras delay between frames and lead to motion blur.
Hence, dynamic context, as complementary information to the static RGB image, can be extracted from event-based data to overcome the drawbacks of intensity images.
These benefits serve as the motivation for applying the emerging technique of event-based neuromorphic vision to autonomous driving in a sensor fusion manner.

Finally, as a preliminary exploration on the accident semantic segmentation task, we propose the \emph{Event-driven Dynamic Context Network (EDCNet)} to capture dynamic context from complicated and volatile driving imagery. For adaptively aggregating the static RGB image and dynamic event data, our EDCNet features \emph{Event Attention Module (EAM)} and \emph{Event Gate Module (EGM)} on the lightweight event branch, which serves as the event-based fusion architecture of the multimodal model, as well as a domain bridge connecting the source (normal) and target (accident) data. In accordance with our EDCNet architecture, the robustness of semantic segmentation towards accident scenarios can be significantly improved, as shown in~\cref{fig:miou_gap}. We benchmark more than $20$ representative accuracy- and efficiency-oriented architectures on our established accident scene segmentation dataset, demonstrating that the proposed EDCNet clearly gets ahead these methods. Furthermore, we have performed extensive experiments by using models learned on multiple source databases including Cityscapes~\cite{cordts2016cityscapes}, KITTI-360~\cite{xie2016semantic}, BDD~\cite{yu2020bdd100k}, and ApolloScape~\cite{wang2019apolloscape}, showing that our proposal is consistently effective for enhancing the reliability of semantic segmentation in accident scenes.

In summary, our main contributions are:
\begin{itemize}
    \item We present a rarely addressed task, \textit{i.e.} \emph{accident scene segmentation} with an accompanying dataset \emph{DADA-seg}, aiming to enhance road safety by improving the robustness of perception algorithms against abnormalities during highly dynamic driving.
    \item We propose \emph{EDCNet}, integrated with \emph{EAM} and \emph{EGM} modules, for exploring event-driven dynamic context. According to the event-aware fusion, comprehensive comparisons and ablation studies are conducted between various datasets for analyzing the benefits and drawbacks of event-based data.
    \item We verify the \emph{EGM} module in a cross-modal event-aware domain adaptation model and conduct a comparison with recent UDA methods. The domain adaptation experiment demonstrates that the foreground segmentation can benefit from the domain-consistent event-based data.
\end{itemize}

It is noteworthy that this work is the extension of our earlier conference paper~\cite{zhang2021issafe}. The extended contents include: 
\begin{itemize}
    \item The performance gap comparison is expanded to include more than $20$ latest CNN- and Transformer-based models.
    \item The event data representation is extensively analyzed by adding an extra time bin setting and an ablation study on the event intensity in data preprocessing.
    \item A comprehensive analysis of the trade-off between accuracy, runtime, and model parameters is advanced.
    \item A new set of quantitative comparisons with recent domain adaptation methods is conducted.
    \item Other enhanced parts are related work reviews, more detailed qualitative analysis, and additional failure analysis.
\end{itemize}

\section{Related Works}
\subsection{Traffic Accident Sensing}
Accompanied by advancing ITS, some existing works tend to explore the edge cases of road driving, aiming to improve the sensing ability on accident scenarios. For identifying the dangerous or safe driving scenes, a binary classifier was boosted by a mixture of experts~\cite{pashaei2020accident_classify}. To further subdivide the categories, the potential danger scene was distinguished by a ternary classifier~\cite{Taccari2018_Near-Crash}. 
Compared with these tasks, we advocate the semantic segmentation task on traffic accident scenarios, which can provide a dense understanding of the driving scene by segmenting most involved objects.

For forecasting the upcoming driving situation in the near future, a few previous works focused on anticipating the traffic accident. Starting from the anticipation task definition~\cite{chan2016anticipating}, an adaptive loss function~\cite{suzuki2018anticipating_loss} was proposed for improving the accuracy of accident anticipation. Different from their bounding box annotations with only four object categories, our proposed dataset has pixel-wise annotations of all $19$ object classes. Bao~\textit{et al.}~\cite{bao2021drive} presented a deep reinforcement learning method with anticipation reward and driver fixation reward, which enabled accident prediction on dashcam videos like the DADA dataset~\cite{FangDADA}. In contrast, we focus on robustifying semantic segmentation in accident scenes. Recently, a video dataset~\cite{you2020traffic_accident_benchmark} for causality in traffic accident scenes with semantic labels of cause and effect, and their temporal intervals, were introduced for the traffic video classification and localization task. For the video analysis tasks, the higher time resolution of the event data explored in this work has a large potential for more fine-grained anticipation of traffic accidents, as well as the temporal classification and localization. 

\subsection{Semantic Segmentation}
Convolutional Neural Networks (CNNs) have dominated the visual understanding field. Since FCN~\cite{long2015FCN} used fully convolutional layers for pixel-wise prediction on images, a massive number of models~\cite{zhao2017PSPNet}\cite{chen2018deeplabv3plus}\cite{yuan2019OCRNet}\cite{fu2019dual}\cite{yin2020dnl} have achieved remarkable performance on image semantic segmentation. Among these accuracy-oriented networks, PSPNet~\cite{zhao2017PSPNet} and DeepLabV3+~\cite{chen2018deeplabv3plus} leveraged multi-scale feature representations.
To harvest contextual information, long-range dependencies~\cite{yuan2019OCRNet}\cite{fu2019dual} were widely explored. DNL~\cite{yin2020dnl} designed a disentangled non-local block to effectively capture global context. ResNeSt~\cite{zhang2020resnest} stacked Split-attention blocks in ResNet style, which greatly benefits downstream semantic segmentation systems. {Recently, vision transformers have also been directly employed for dense image segmentation~\cite{setr}\cite{segformer}.} Moreover, unifying semantic- and instance-specific segmentation, current panoptic segmentation provided by a single model like Panoptic FPN~\cite{Kirillov2019PanopticFPN} has achieved excellent results for its sub-tasks and upper-level navigation applications.
In addition to high accuracy, other works, such as ERFNet~\cite{romera2017erfnet}, SwiftNet~\cite{orsic2019swiftnet}, and Fast-SCNN~\cite{poudel2019fastscnn}, proposed lightweight architectures to improve the efficiency. In specific, ERFNet leveraged early downsampling and filter factorization. SwiftNet designed ladder-style upsampling, {whereas} Fast-SCNN followed a multi-branch setup. Furthermore, compact classification networks like MobileNet~\cite{Howard_2019MobileNetV3} were frequently applied to accelerate semantic segmentation. 
In the ITS field, sEnDec~\cite{akilan2019sendec} was proposed to achieve more delineated foreground segmentation even in adverse conditions like bad weather and night videos. An end-to-end multi-frame framework was designed in~\cite{patil2020end} for moving object segmentation, which embeds edge- and motion features and has been examined on weather-degraded traffic scenes.
In this work, we explore dynamic context by mining event-driven semantics like fast moving objects to boost the segmentation performance.

Most of the above mentioned methods are designed for normal conditions on major segmentation benchmarks~\cite{cordts2016cityscapes}\cite{xie2016semantic}\cite{yu2020bdd100k}\cite{wang2019apolloscape}. In spite of the improved accuracy and efficiency, the learned understanding does not adequately generalize well to new scenes such as adverse weathers, which invites Domain Adaptation (DA) strategies, aiming to adapt segmentation models to previously unseen domains. {Wang~\textit{et al.}~\cite{wang2020differential} proposed a differential treatment approach for stuff and thing classes when performing domain alignment. Chen~\textit{et al.}~\cite{chen2020naive} presented a semi-supervised learning method to make use of video sequence data for creating pseudo labels.}

To list a few {of DA methods for adverse conditions}, the day-night conversions in~\cite{song2020nighttime} and the adaptations between diverse weathers like foggy~\cite{sakaridis2018modelAdapt_foggy} and rainy~\cite{di2020rainy} environments.
However, apart from these natural conditions in real driving scenes, there are many uncontrollable factors in the interaction with other traffic participants. The core purpose of our work is to fill the gap of semantic segmentation in abnormal situations.

Any ambiguity in machine vision algorithms may cause fatal consequences in autonomous driving~\cite{liu2020wasserstein}, thus robustness testing in diverse driving conditions is essential. For this reason, WildDash~\cite{zendel2018wilddash} provided ten different hazards, such as blurs, underexposures or lens distortions, as well as negative test cases against the overreaction of segmentation algorithms. Fishyscapes~\cite{blum2019fishyscapes} evaluated pixel-wise uncertainty estimations towards the segmentation of anomalous objects. MSeg~\cite{lambert2020mseg} constructed a composite dataset with the aim of facilitating generalization for multi-domain semantic segmentation.
Inspired by these works, we create a new dataset to extend the robustness test from ordinary to accident scenarios. In our DADA-seg dataset, most of the critical or accidental scenes are more difficult by having a large variety of adverse hazards.

On the other hand of improving robustness, some solutions constructed a multimodal segmentation model by fusing additional information, such as depth information in RFNet~\cite{sun2020rfnet}, thermal information in RTFNet~\cite{Y2019RTFNet_rgb_thermal}, and optical flow in~\cite{rashed2019optical_flow}.
Differing from these classic modalities, event-based data will be explored as a novel auxiliary modality. Specifically, we propose an attention-bridged architecture to extract dynamic context and thereby improve accident scene segmentation.

\subsection{Event-based Vision}
Event cameras, such as the Dynamic Vision Sensor (DVS)~\cite{patrick2008DVS_sensor} and Asynchronous Time-based Image Sensor (ATIS),
are novel perceptual sensors inspired by biology. Its working principle is completely different from traditional frame-based cameras. Whereas the typical camera indistinguishably captures the global scene as an image at a fixed frame rate, the event camera is triggered by dynamic change of intensity at individual pixel in the scene. Compared to normal cameras, the event camera has some complementary characteristics, such as high dynamic range ($140 dB$), no motion blur, and response in microseconds~\cite{gallego2019event_survey}. Recently, different approaches based on event data have been proposed, such as 3D reconstruction and 6-DOF tracking~\cite{kim2016_3drecon_event}, monocular depth prediction~\cite{gehrig2021combining}, optical flow estimation~\cite{zhu2018ev-flownet}, as well as object detection and recognition~\cite{mitrokhin2018objdetect_event}. 
Event cameras asynchronously encode intensity changes at each pixel with position, time, and polarity: $(x, y, t, p)$. For processing the sparse event data in a deep CNN, the original spatial-temporal event stream is often converted into an image form by remaining the higher time resolution, such as a two-channel event frame by Maqueda~\textit{et al.}~\cite{maqueda2018eventframe_2channel}, a four-dimensional grid~\cite{zhu2018ev-flownet}, and a Discretized Event Volume (DEV) in x-y-t space by Zhu~\textit{et al.}~\cite{zhu2019unsupervised}.

According to the image-like transformation, Alonso~\textit{et al.}~\cite{alonso2019Ev-SegNet} introduced a six-channels event representation and constructed a semantic segmentation model Ev-SegNet on an extended event dataset DDD17~\cite{binas2017ddd17}, whose semantic labels are generated by a pre-trained model on Cityscapes and only contain $6$ major categories. In contrast, our models are trained with the ground-truth labels of Cityscapes and perform semantic segmentation in all $19$ object classes, so that the perception component can deliver a sufficiently dense and fine-grained scene understanding result for upper-level assistance systems. Additionally, instead of stacking RGB images with the event frames in the input stage, the event data will be adaptively fused with the RGB image through the proposed attention mechanisms, which are more effective for combining the two heterogeneous modalities.

However, for event-based vision, labeled event data for semantic segmentation is scarce in the state of the art. Previous works leveraged the existing labeled data of images by simulating their corresponding event data. Rebecq~\textit{et al.}~\cite{rebecq2018esim} proposed ESIM to combine a rendering engine with an event simulator, which allows the event simulator to adaptively query visual frames based on the dynamics of the visual signal. In~\cite{gehrig2020video2event}, event generation and frame upsampling methods were incorporated to recycle the existing video datasets for training networks designed for real event data. Recently, without rendering engine nor upsampling, EventGAN~\cite{zhu2019eventgan} presented a self-supervised approach and constructed a GAN model to directly generate event frames from the associated images using only modern GPUs. 

In order to guarantee the accuracy of training labels, we primarily select those datasets~\cite{cordts2016cityscapes}\cite{xie2016semantic}\cite{yu2020bdd100k}\cite{wang2019apolloscape} with manual annotations as the source domain, which also maintain the category consistency with our proposed accident dataset \emph{DADA-seg} as the target domain. To extend the source and target datasets, we utilize the EventGAN model to generate their associated event data.

\section{Methodology}
In this section, we state the details of the accident semantic segmentation task and the relevant dataset, as well as our \emph{EDCNet} architecture, attempting to tackle the performance drop of image semantic segmentation algorithms in accident scenes by capturing event-driven dynamic context.

\subsection{Task Definition}\label{method:task_definition}
To bring IV systems closer to real applications, we focus on \emph{accident scene segmentation}. Besides, an associated evaluation set following the same labeling rules as Cityscapes~\cite{cordts2016cityscapes} is provided for quantitative comparison and analysis. All test cases are collected from real-world traffic accidents and most contain adverse factors. We explicitly study the robustness in challenging accident scenarios based on the assumption that the less performance degradation of the algorithm in this unseen dataset, the better its robustness. 

\begin{table}[]
\centering
\caption{\small Distribution of total $313$ sequences from DADA-seg dataset under conditions in terms of light, weather, and occasion.}
\label{tab:condition_distribute}
\begin{tabular}{c|c@{\hskip 3pt}c|c@{\hskip 3pt}c|c@{\hskip 3pt}c@{\hskip 3pt}c@{\hskip 3pt}c@{\hskip 3pt}}
\toprule
\multirow{2}{*}{\textbf{DADA-seg}}   & \multicolumn{2}{c|}{\textbf{Light}} & \multicolumn{2}{c|}{\textbf{Weather}} & \multicolumn{4}{c}{\textbf{Occasion}}     \\ \cmidrule{2-9}
           & day & night & sunny & rainy & highway & urban & rural & tunnel \\ \midrule \midrule
\#sequence & 285 & 28    & 297   & 16    & 32      & 241   & 38    & 2      \\ \bottomrule
\end{tabular}
\vskip -3ex
\end{table}

\subsection{Accident Dataset: DADA-seg}\label{method:dataset}
\textbf{Data Annotation.} Our proposed dataset DADA-seg is selected from the DADA-2000~\cite{FangDADA} dataset, which was collected from various mainstream video sites, such as Youtube, Youku, Bilibili, iQiyi, Tencent, \textit{etc.} It was proposed for driver attention prediction and consists of $2,000$ sequences with the resolution of $1,584{\times}660$. However, it includes only the accident category labels and the attention maps, which are not used in this work. In order to extend to the accident semantic segmentation task, we performed additional laboratory preprocessing work based on all $2,000$ sequences in two stages. In the first stage, we cleaned the whole dataset by removing some sequences in the case with: \textit{(i)} large watermarking; and \textit{(ii)} low resolution. Meanwhile, most of the sequences with typical adverse factors were retained, such as those with motion blur, over/underexposure, weak illumination, occlusion, \textit{etc.} Furthermore, concentrating on accident scenes, we modified the frame distribution of the original accident window, by remaining $10$ frames before the accident and $30$ frames during the accident. After our selection, a total of $313$ sequences containing $12,520$ available frames were retained to constitute our final DADA-seg dataset. 

In the second stage, we manually perform the pixel-wise annotation on every $11$-th frame of each sequence by using the polygons to delineate the fine-grained boundaries for individual semantic objects. The object class definition of the annotation follows the labeling rule of Cityscapes. The annotation samples are shown in the $t_1$ frame in \cref{fig:DADA_SegExample}. Finally, the DADA-seg dataset includes $313$ labeled images for quantitative analysis of accident semantic segmentation and $12,207$ unlabeled images for event-aware domain adaptation between the normal and abnormal imagery.

The distribution is described in Table~\ref{tab:condition_distribute}. Compared with the Cityscapes dataset, all images from our dataset are taken in a broad range of regions by different cameras from various viewpoints, so as to maintain a larger diversity of driving scenes. Thus, the DADA-seg dataset covers the evaluation and verification of models which are trained on and transferred from other source datasets. Besides, all sequences are collected from traffic accident scenarios, comprising normal, critical, and accidental situations, which can be treated as edge cases of road-driving environments.  

\subsection{Event-based Data}\label{method:event data}
\textbf{Event Data Representation.} Event cameras asynchronously encode an event at each individual pixel ($x$, $y$) at the corresponding triggering timestamp $t$, if the change of logarithmic intensity $L(x, y, t) = Log(I(x, y, t))$ in time variance $\Delta t$ is greater than a preset threshold $C$. Eq.~\eqref{eq:1} denotes the condition of triggering an event at $(x,y,t)$.
\begin{equation}\label{eq:1}
    L(x, y, t) - L(x, y, t - \Delta t) \geq  pC, \; p\in \left \{ -1, +1 \right \},
\end{equation}
where polarity $p$ indicates the positive or negative direction of change. As in Eq.~\eqref{eq:2}, a typical volumetric representation of a continuous event stream with size $N$ is a set of $4$-tuples, where each event is represented as an independent tuple consisting of the position $(x,y)$, the timestamp~$t$, and the polarity~$p$.
\begin{equation}\label{eq:2}
    V = \left \{ e_i \right \}_{i=1}^N, \textup{where} \; e_i = (x_i, y_i, t_i, p_i).  
\end{equation}

\begin{figure}[!t]
	\centering
    \includegraphics[width=1.0\linewidth]{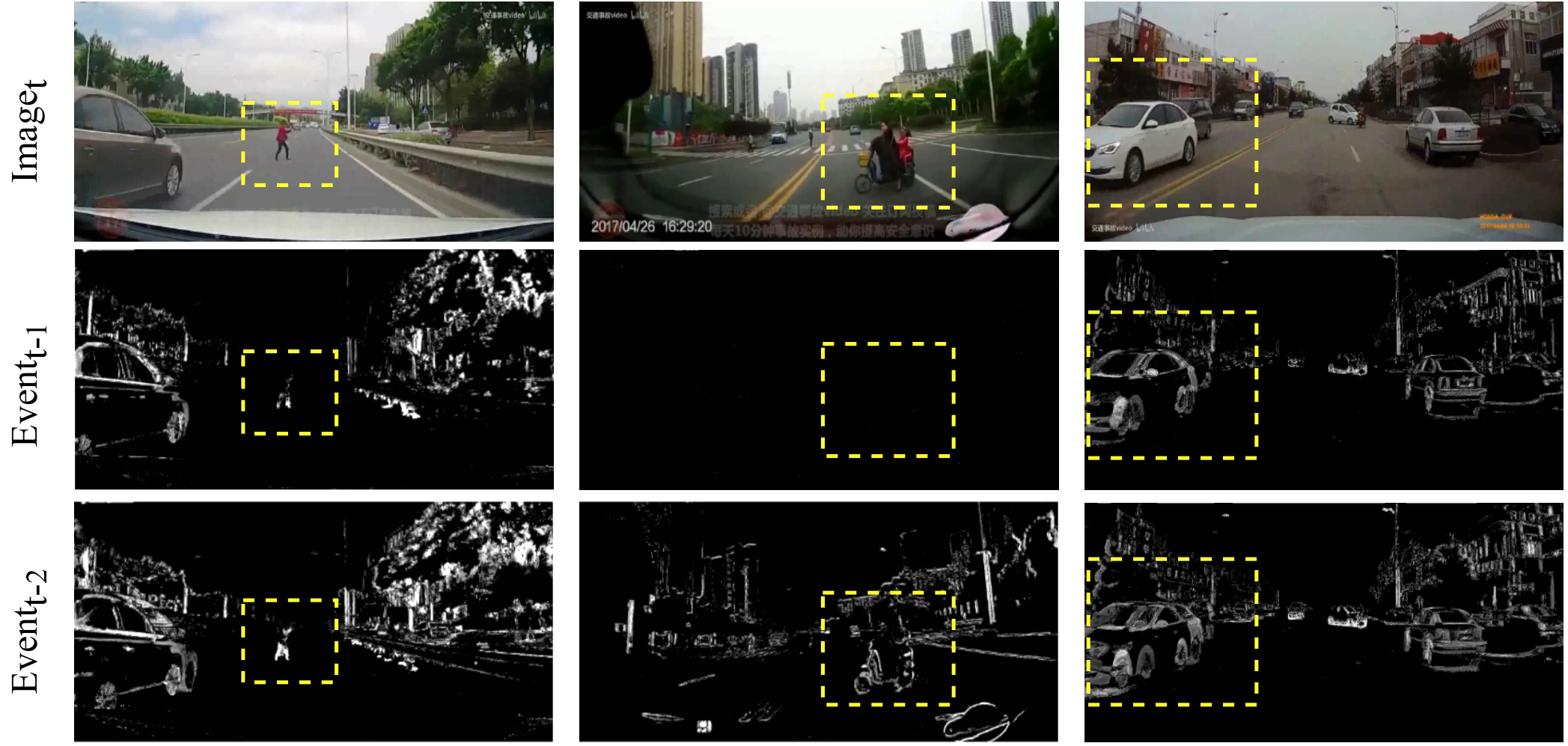}
    \begin{minipage}[t]{.3\linewidth}
    \centering
    \vskip -2ex
    \subcaption{Pedestrian}\label{fig:pedestrian_intensity}
    \end{minipage}%
    \begin{minipage}[t]{.3\linewidth}
    \centering
    \vskip -2ex
    \subcaption{Motorcyclist}\label{fig:motorcyclist_intensity}
    \end{minipage}%
    \begin{minipage}[t]{.3\linewidth}
    \centering
    \vskip -2ex
    \subcaption{Car}\label{fig:tailing_car}
    \end{minipage}%
    \vskip -2ex
	\caption{\small Comparison of event generation on DADA-seg dataset. Given the same anchor frame at time $t$, the $Event_{t-2}$ generated through the frame pair $(t-2, t)$ contains more details than the $Event_{t-1}$ by $(t-1, t)$, especially for the moving foregrounds highlighted by yellow dashed boxes. Zoom in for a better view.}
	\label{fig:dada_generated_event_frame}
\vskip -2ex
\end{figure}

However, it is still arduous to transmit the asynchronous event spike to the convolutional network by retaining a sufficient time resolution. Hence, we perform a dimensionality reduction operation~\cite{zhu2019eventgan} in the time dimension. After the ratio of the original timestamp is calculated by $(t_{i}-t_{1}) /(t_{N}-t_{1})$, the new timestamp $\Tilde{t_{i}}$ of $i$-th event can be calculated via Eq.~\eqref{eq:3}. Then, the original volume $V$ of Eq.~\eqref{eq:2} is discretized with a fixed length for positive and negative events separately. Each event is locally linearly embedded to the nearest time-series panel, similar to the bilinear interpolation. According to the preset number of positive time bin $B^+$, a discretized spatial-temporal volume $V^+$ is represented as Eq.~\eqref{eq:4}.
\begin{equation}\label{eq:3}
    \Tilde{t_{i}} =(B^+-1)\left(t_{i}-t_{1}\right) /\left(t_{N}-t_{1}\right).
    \vspace{-0.2cm}
\end{equation}
\begin{equation}\label{eq:4}
    V^+(x, y, \Tilde{t_{i}}) =\sum_{i}^{B^+} \max \left(0,1-\left|t-\Tilde{t_{i}}\right|\right).
\end{equation}
After the dimensionality reduction for discretization of the respective positive and negative time bins, both positive and negative sub-volumes are concatenated along the time dimension to construct the whole volume $V \in \mathbb{R} ^{B \times W\times H}$, where $B=B^+ + B^-$, and $B$ is the total number of time bins, while $W$ and $H$ are the width and height of spatial resolution, respectively. Based on the reconstructed event volume in Eq.~\eqref{eq:4}, the high resolution temporal information of the event data can be maintained, which is a vital factor for extracting dynamic features. The detailed setting of time bins $B$ will be discussed in the experiments section.

\begin{figure}[!t]
	\centering
    \includegraphics[width=0.95\linewidth]{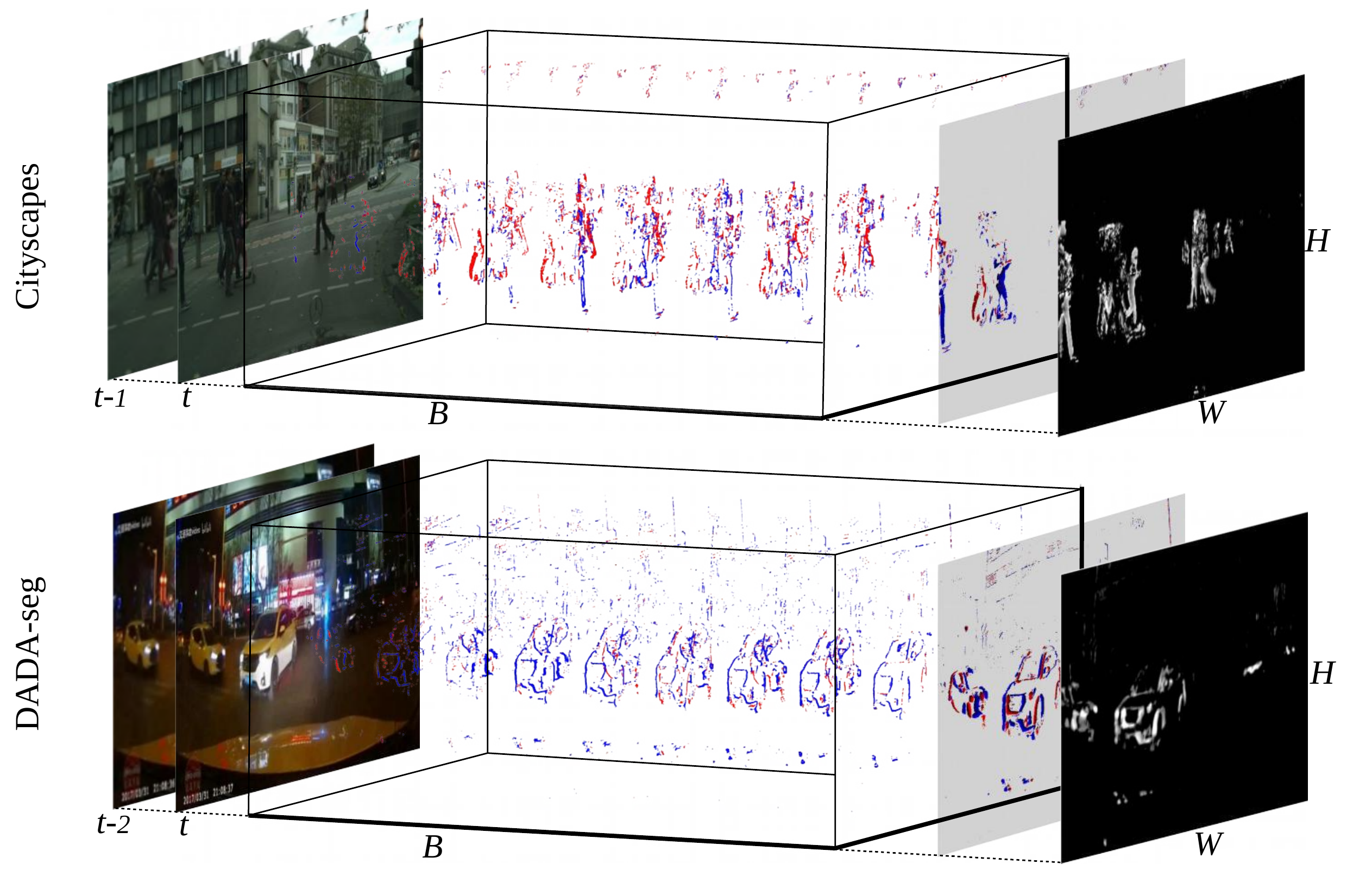}
	\vskip -1ex
	\caption[]{\small Visualization of generated event data in $B\times H \times W$ space, where $B$, $H$ and $W$ denote the time bins, image height and width. From left to right are RGB image pair ($I_{t-1}$, $I_{t}$), different event representations according the time bins number: event volume with higher time resolutions, event polarity frame and event grayscale frame, where blue and red colors indicate positive and negative events, respectively.} 
	\label{fig:event_gen}
\vskip -3ex
\end{figure}

\begin{table*}[t]
\centering
\caption{Statistics of datasets after event data synthesis. Merge3 dataset is combined with the Cityscapes, KITTI-360, and BDD3K, having identical label mapping. DADA-seg dataset serves as a target domain and has no training data.}
\label{tab:stat_datasets}
\begin{tabular}{l|r|r|r|r|r|r}
\toprule
\textbf{Datasets} & \textbf{Cityscapes} & \textbf{KITTI-360} & \textbf{BDD3K} & \textbf{ApolloScape} & \textbf{Merge3} & \textbf{DADA-seg} \\ \midrule
\#training & 2,975 & 5,504 & 3,086 & 6,056 & 11,565 & 0 \\ 
\#evaluaion & 500 & 612 & 343 & 673 & 14,555 & 313 \\ \bottomrule
\end{tabular}
\vskip -4ex
\end{table*}

\textbf{Event Data Synthesis.} Due to the lack of event-based labeled dataset for the semantic segmentation task, we consider synthesizing event data from RGB images that have manually labeled annotations, instead of generating insufficiently accurate pseudo labels based on grayscale images that have real event data. Hence, we utilize the EventGAN~\cite{zhu2019eventgan} model that was trained from the MVSEC dataset~\cite{zhu2018MVSEC} including around $30$-minute real event data and image sequences, to synthesize highly reliable event data on different datasets~\cite{cordts2016cityscapes}\cite{xie2016semantic}\cite{yu2020bdd100k}\cite{wang2019apolloscape} for the semantic segmentation task. Different from the fixed frame rate ($17Hz$) in the Cityscapes dataset~\cite{cordts2016cityscapes}, the sequence in our DADA-seg dataset was acquired with diverse cameras and frame rates, which indicates that its synthesized event data vary from the intensity of motion due to different time intervals. {To discover a suitable intensity of simulated event data for DADA-seg, we verify different frame intervals in ~\cref{fig:dada_generated_event_frame}. After qualitative analysis and considering the benefit to obtain more event intensity (see Fig.~\ref{fig:motorcyclist_intensity}) meanwhile having less tailing issues (see Fig.~\ref{fig:tailing_car}), we select the penultimate frame and stack with its anchor frame that has the semantic annotation, to form as the image pair for event data synthesis.} Two cases of the simulated event data are visualized in~\cref{fig:event_gen} under different temporal dimensions. From this, it can be easily noticed how event data benefits sensing in the dynamic driving scene with moving objects or in the low-lighting environment, meanwhile providing higher time resolution in the volumetric form.

In order to evaluate our proposed EDCNet more comprehensively, thorough comparison experiments in different semantic segmentation datasets are conducted, which will be unfolded in the following section. Since the synthesis of event data requires associated image pairs ($I_{t-1}$, $I_{t}$), where $I_{t}$ is the anchor image with semantic annotation, the previous image with respect to the anchor image must be available in the dataset as the same time. Thus, two conditions should be met to select a dataset: \textit{(i)} the anchor image has its semantic annotation for model training; and \textit{(ii)} the image sequence is available.

\begin{figure*}[!t]
	\centering
    \includegraphics[width=0.99\linewidth]{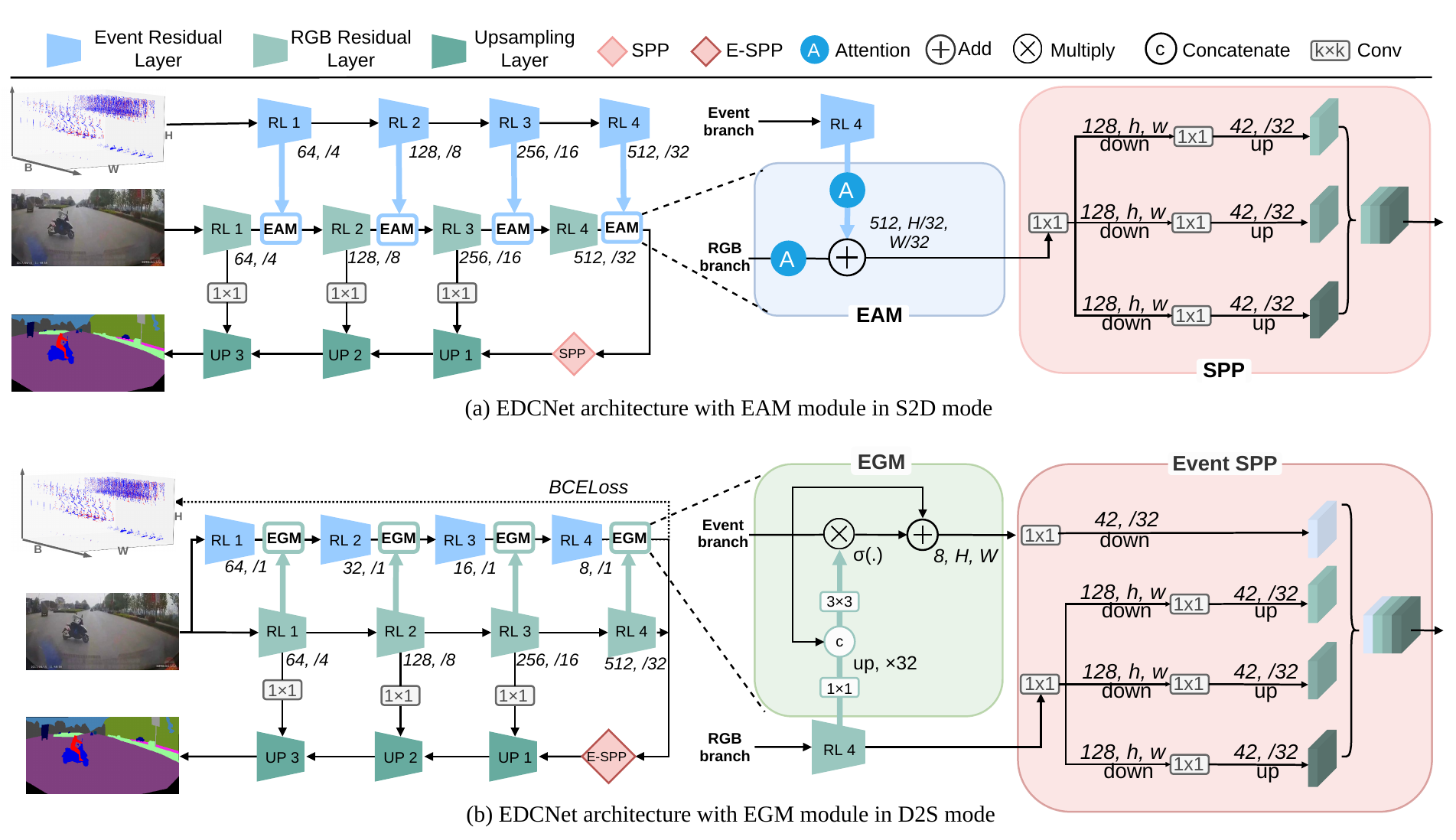}
	\caption{Model architectures of two different event fusion strategies. In (a) S2D mode with EAM module, event data are fused to RGB branch adaptively from sparse to dense, while in (b) D2S mode with EGM module, event data are extracted from dense image and learned from the sparse ground truth.} 
	\label{fig:event_fuse}
\vskip -1ex
\end{figure*}

Finally, in addition to the Cityscapes dataset with $2,975$ training and $500$ validation images, we selected three datasets that meet the above conditions for our study, which are KITTI-360~\cite{xie2016semantic}, BDD~\cite{yu2020bdd100k}, and ApolloScape~\cite{wang2019apolloscape}. The ApolloScape and KITTI-360 datasets have semantic annotations for each frame of their video sequences. Therefore, we only sample one anchor image every $10$ frames from the video sequence to prevent overfitting cases. After sampling, the KITTI-360 dataset has $5,504$ training and $612$ validation images, while the ApolloScape dataset has $6,056$ training and $673$ validation images. However, due to the limited amount of anchor images having annotations in the BDD dataset, only $3,086$ training and $343$ validation images were filtered out based on the aforementioned two conditions, which is termed as \textit{BDD3K} in our work. It should be noted that the category definition of the ApolloScape dataset is different from other datasets, so those models trained on it perform segmentation with only $16$ overlapping categories~\cite{wang2019apolloscape}. Based on the selected RGB image pairs from all datasets, the same method described above is utilized to synthesize the associated event data, and the procedure will be open-sourced to foster future research. {Statistics of the datasets are listed in Table~\ref{tab:stat_datasets}}.

\subsection{EDC: Event-driven Dynamic Context}

For capturing the event-driven dynamic context to enhance the feature extracted from static RGB image we propose the \emph{Event-driven Dynamic Context Network (EDCNet)}. The entire architecture of EDCNet is shown in~\cref{fig:event_fuse}. Motivated by the high time resolution characteristic of event data, the sparsity of the original event data should be preserved in the process of integrating events, so as to obtain finer-grained motion features from event frames. But in turn, due to the sparsity and dynamics of event data compared to RGB images, the straightforward manners~\cite{zhu2018ev-flownet}\cite{maqueda2018eventframe_2channel}\cite{alonso2019Ev-SegNet} by stacking them in the early stage are {insufficient} to aggregate those two heterogeneous modalities. Thereby, we construct \emph{Event Attention Module (EAM)} and \emph{Event Gate Module (EGM)} based on attention mechanisms to enhance the fusion process and apply them in different stages of our EDCNet to ensure that these modalities can be aligned effectively and integrated adaptively. 

With respect to different data characteristics, we mainly explore two fusion modes, \textit{i.e.} \emph{Sparse-to-Dense mode (S2D)} and \emph{Dense-to-Sparse mode (D2S)}. Correspondingly, the EAM module is applied to the EDCNet in the S2D mode, {whereas} the EGM module is applied in the D2S mode.

\textbf{Event Attention Module.} 
An intuitive sparse-to-dense process mode is capable of treating the sparse event data as input and extracting dense features. Inspired by the design of SwiftNet~\cite{orsic2019swiftnet} and RFNet~\cite{sun2020rfnet}, our initial EDCNet in the S2D mode with EAM module extends the multimodal architecture with more sparse event data and an adapted EAM module. In our model, the original fusion mechanism is modified from dense depth information to sparse event-based data, so as to support the event-aware fusion in the proposed S2D mode. As shown in~\cref{fig:event_fuse}a, the EAM module is constructed with channel-wise attention block~\cite{hu2018SEBlock} for feature selection, which can harvest motion features from the original event data. In the S2D design, while the four residual blocks in the RGB branch gradually extract higher-level features at smaller resolutions with \{$4$, $8$, $16$, $32$\} downsampling rates and \{$64$, $128$, $256$, $512$\} channels, the accompanying Event branch also processes event data in the same manner. The EAM module is applied between each two blocks corresponding to the RGB and Event branches, so as to adaptively extract motion-related features from dynamic event data. 
Given the image feature map termed as $\mathcal{F}_{i} \in \mathbb{R} ^{C \times H\times W}$ and the event feature map as $\mathcal{F}_{e} \in \mathbb{R} ^{\hat{C} \times H\times W}$, EAM module adaptively merges them to form a new feature map termed as $\mathcal{F}_{EAM} \in \mathbb{R} ^{C \times H\times W}$. The calculation inside EAM module is represented as Eq.~\eqref{eq:5}.
\begin{equation}\label{eq:5}
    \mathcal{F}_{EAM}=\mathcal{F}_{i} \otimes \sigma_i\left[f(\mathcal{F}_{i})\right] + \mathcal{F}_{e} \otimes \sigma_e\left[g(\mathcal{F}_{e})\right],
\end{equation}
where both $f(\cdot)$ and $g(\cdot)$ are composed of adaptive global pooling and $1{\times}1$ convolution operations, the $\sigma_i(\cdot)$ and $\sigma_e(\cdot)$ denote Sigmoid activate function for the image and event feature map. Benefiting from the channel-wise attended EAM module, the event data with higher time resolution compensate for dynamic context and provide motion-related features {insufficient} in static images, so as to improve the representational capacity of the merged feature map $\mathcal{F}_{EAM}$. Additionally, its high dynamic range enhances over-/underexposure images. After four residual blocks, the merged feature with a shape of $512 \times \frac{H}{m} \times \frac{W}{m}$ will be the straightway input to the Spatial Pyramid Pooling (SPP) module~\cite{zhao2017PSPNet} for capturing multi-scale features, where the $m$ is the downsampling rate. In this S2D mode, we build the SPP module via multi-scale grids in three different resolutions of $(h, w) \in \{(8, 16), (4, 8), (2, 4)\}$. These grid features will be uniformly interpolated to the same original resolution, and be concatenated as a new feature map, as shown in the right side of \cref{fig:event_fuse}. Afterwards, a lightweight decoder, composing of three upsampling modules with blend convolution operations, will perform element-wise addition to fuse the upsampled feature map with $1{\times}1$ skip connections from the RGB branch, so as to align different levels of feature maps for the final prediction. 

\textbf{Event Gate Module.}
Aside from the intuitive S2D mode, inspired by the video restoration from a single blurred image and the event data like~\cite{pan2019video_from_event}\cite{jin2018video_from_blurred_image}, we alternatively leverage the dense-to-sparse~(D2S) mode, aiming to capture sparse event information and dynamic context. In the D2S design, the event branch can firstly benefit from the identical dense feature dividing from the RGB branch. Besides, through the supervised learning with the ground-truth event data, the event branch can further guarantee the effectiveness of the dynamic context. For constructing our EDCNet in the D2S mode, we perform three major modifications compared to the previous S2D mode, as illustrated in ~\cref{fig:event_fuse}b. Firstly, the standard event residual layers with multiple convolution blocks are replaced by a single block, since such a shallow event branch is more effective to extract features from dense to sparse and it is capable of processing the event feature at a higher spatial resolution without overloading the computation demand. In detail, after the $7{\times}7$ stem block of ResNet-18~\cite{he2016resnet}, while the RGB branch encodes higher-level features at smaller resolutions with \{$4$, $8$, $16$, $32$\} downsampling rates and \{$64$, $128$, $256$, $512$\} channels, the event branch gradually shallows event channels in the order of \{$64$, $32$, $16$, $8$\}, which also enables event processing at a full resolution. Secondly, we propose a novel gating mechanism named Event Gate Module (EGM), which is capable of aggregating the dense RGB feature map $\mathcal{F}_{i} \in \mathbb{R} ^{C \times H\times W}$ and the sparse event feature map $\mathcal{F}_{e} \in \mathbb{R} ^{\hat{C} \times H\times W}$, and forming a new event-aware feature map $\mathcal{F}_{EGM} \in \mathbb{R} ^{\hat{C} \times H\times W}$. As denoted in Eq.~\eqref{eq:6}, the image and event feature maps are fused by the internal gating process of the EGM module.
\begin{equation}\label{eq:6}
    \mathcal{F}_{EGM}=\mathcal{F}_{e} \otimes \sigma\left[c(\mathcal{F}_{e}, g(\mathcal{F}_{i}))\right] + \mathcal{F}_{e},
\end{equation}
{where $g(\cdot)$ is a $1{\times}1$ convolution, $\sigma(\cdot)$ and $c(\cdot)$ denote Sigmoid activate function and concatenation operations, respectively.} Based on the carefully designed EGM module, we enforce the event branch to deactivate the non-event features according to the higher-level semantic features from the RGB branch. Last but not least, we build the \emph{Event-aware Spatial Pyramid Pooling (E-SPP)} block by adding an event stream in the aforementioned SPP module. This crucial modification of the context extraction module (E-SPP) allows our EDCNet to integrate significant dynamic context and multi-scale features in high-level dimensions. Due to the two-polarity representation of event data, we select the standard Binary Cross Entropy (BCE) to form the event loss function $\mathcal{L}_{event}$, while the Cross Entropy (CE) is selected for semantic segmentation. The event-related supervision will be performed on the event branch before the E-SPP module. Aiming to learn the whole model in an end-to-end fashion, the final loss function is constructed as in Eq.~\eqref{eq:7}, in which the event loss function $\mathcal{L}_{event}$ is added to the segmentation loss function $\mathcal{L}_{seg}$.
\begin{equation}\label{eq:7}
\mathcal{L}=\mathcal{L}_{event}(e, \hat{e}) + \mathcal{L}_{seg}(y, \hat{y}),
\end{equation}
where $e$, $\hat{e}$, $y$ and $\hat{y}$ are the ground-truth and the predicted event, segmentation ground truth and prediction, respectively.

\begin{figure*}[!t]
    \centering
    \includegraphics[width=0.9\textwidth]{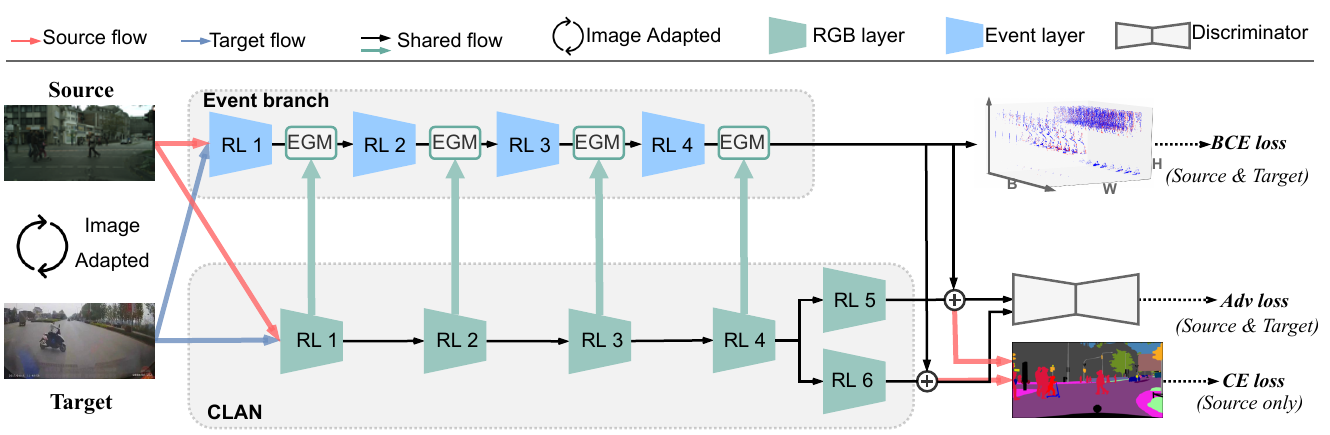}
    \vskip -1ex
    \caption{Architecture of EDCNet with the cross-modal event-aware domain adaptation strategy. The Event branch is constructed by the EGM module in the D2S fusion mode.}
    \label{fig:event_fuse_CLAN_D2S}
\vskip -3ex
\end{figure*}

\subsection{EDA: Event-aware Domain Adaptation}
Initially, due to the scarcity of training data concerning accidental driving scene, we tend to apply Unsupervised Domain Adaptation (UDA) to perform the transfer learning from normal to abnormal dataset. For exploring potential from the large number of unlabeled samples in our dataset, we investigate the UDA strategy in different aspects, \textit{i.e.} image level~\cite{zhu2017cycleGAN} and/or feature level~\cite{luo2019CLAN}. More importantly, compared to textured RGB images, the monochromatic event data, capturing only changes of intensity, is semantically more consistent in both domains. It denotes that the homogeneous event features can serve as a bridge to assist the RGB modal domain adaptation in the feature level as well. Based on this assumption, as shown in~\cref{fig:event_fuse_CLAN_D2S}, we construct our EDCNet in an \emph{Event-aware Domain Adaptation (EDA)} strategy with two branches, where the lightweight event branch is the same as the aforementioned D2S mode and the RGB branch is created by the ResNet-101 backbone~\cite{he2016resnet} referring to the CLAN model~\cite{luo2019CLAN}. Summarily, in the EDA strategy, we aim to boost the performance of EDCNet by using: \textit{(i)} unlabeled accident RGB images, \textit{(ii)} monochromatic event data of two domains, and \textit{(iii)} a larger backbone. Up to our knowledge, we are making an early attempt to perform cross-modal unsupervised domain adaptation from normal to abnormal driving scenes between two heterogeneous modalities. In order to maintain the consistency of ground-truth labels of both branches, in the corresponding experiment, we mainly discuss the D2S mode, from which the original event data is applied as supervision signals instead of as inputs.

To distinguish and {investigate} the impact of diverse domain adaptation strategies, we utilize the CycleGAN model~\cite{zhu2017cycleGAN} to translate style of images from Cityscapes to DADA-seg and perform image-level adaptation between the two domains. 

\section{Experiments}
In this section, comprehensive experiments in different settings and implementation details are presented. Initially, performance gaps of various semantic segmentation models are investigated. Afterwards, comprehensive experiments verify the effectiveness of the proposed \emph{EDCNet} in reducing the performance gap. The analysis of the event-aware domain adaptation (EDA) strategy is also conducted.

\begin{table}[t]
\caption{\small Performance gap of models, which are trained and validated on Cityscapes and then tested on DADA-seg, both with a $1024{\times}512$ resolution. {Transformer-based models in the fourth group are tested in $768{\times}768$ resolution following their default settings.}}
\label{tab:domain_gap}
\centering
\resizebox{\columnwidth}{!}{
\begin{tabular}{@{}llrrc@{}}
\toprule
\textbf{Network} & \textbf{Backbone} & \textbf{Cityscapes} & \textbf{DADA-seg} & \textbf{mIoU Gap} \\ \midrule \midrule
ERFNet~\cite{romera2017erfnet}          & ERFNet    & 72.1  & 9.0   & -63.1 \\ 
PSPNet~\cite{zhao2017PSPNet}            & MobileNetV2  & 70.2  & 17.1  & -52.5 \\
DeepLabV3+~\cite{chen2018deeplabv3plus} & MobileNetV2  & 75.2  & 16.5  & -58.7 \\ 
MobileNetV3~\cite{Howard_2019MobileNetV3} & MobileNetV3small & 64.1 & 18.2 & -45.9 \\
MobileNetV3~\cite{Howard_2019MobileNetV3} & MobileNetV3 & 69.5 & 17.6  & -51.9 \\
SwiftNet~\cite{orsic2019swiftnet}       & ResNet-18    & 75.4  & 20.5  & -54.9 \\
DeepLabV3+~\cite{chen2018deeplabv3plus} & ResNet-18    & 76.8  & 19.1  & -57.7 \\ 
OCRNet~\cite{yuan2019OCRNet}            & HRNetV2p-W18small & 77.1 & 20.5  & -56.6 \\
Fast-SCNN~\cite{poudel2019fastscnn}     & Fast-SCNN	   & 69.1  & 26.3  & -42.8 \\
\midrule
DeepLabV3+~\cite{chen2018deeplabv3plus} & ResNet-50    & 80.1  & 18.0  & -62.1 \\ 
PSPNet~\cite{zhao2017PSPNet}            & ResNet-50    & 78.6  & 16.1  & -62.5 \\
DANet~\cite{fu2019dual}                 & ResNet-50    & 79.3  & 17.6  & -61.7 \\
DNL~\cite{yin2020dnl}                   & ResNet-50    & 79.3  & 15.7  & -63.6 \\
Semantic FPN~\cite{Kirillov2019PanopticFPN} & ResNet-50 & 74.5 & 18.2  & -56.3 \\
OCRNet~\cite{yuan2019OCRNet}            & HRNetV2p-W18 & 78.6  & 23.0  & -55.6 \\ 
\midrule
DeepLabV3+~\cite{chen2018deeplabv3plus} & ResNet-101   & 80.9  & 21.0  & -59.9 \\
PSPNet~\cite{zhao2017PSPNet}            & ResNet-101   & 79.8  & 20.1  & -59.7 \\
DANet~\cite{fu2019dual}                 & ResNet-101   & 80.4  & 17.8  & -62.6 \\
DNL~\cite{yin2020dnl}                   & ResNet-101   & 80.4  & 19.7  & -60.7 \\
Semantic FPN~\cite{Kirillov2019PanopticFPN} & ResNet-101 & 75.8 & 19.6 & -56.2 \\
ResNeSt~\cite{zhang2020resnest}         & ResNeSt-101  & 79.6  & 20.0  & -59.6 \\
OCRNet~\cite{yuan2019OCRNet}            & HRNetV2p-W48 & 80.7  & 24.5  & -56.2 \\ 
\midrule[1px]
SETR-Naive~\cite{setr} & Transformer-Large & 77.9 & 27.1 & -50.8 \\
SETR-MLA~\cite{setr} & Transformer-Large & 77.2 & 30.4 & -46.8 \\
SETR-PUP~\cite{setr} & Transformer-Large & 79.3 & 31.8 & -47.5 \\
SegFormer-B1~\cite{segformer} & MiT-B1 & 78.0 & 16.6 & -61.4 \\
SegFormer-B2~\cite{segformer} & MiT-B2 & 80.5 & 21.2 & -59.3 \\
SegFormer-B3~\cite{segformer} & MiT-B3 & 81.5 & 27.0 & -54.5 \\
\bottomrule
\end{tabular}}
\vskip-3ex
\end{table}

\begin{table*}[!t]
\setlength{\tabcolsep}{10pt}
\renewcommand{\arraystretch}{0.9}
\caption{Comparison of different event representations and event fusion approaches. All models use ResNet-18 as backbone and are tested with a $1024{\times}512$ resolution. \textit{B} is short for the time bins of event data. The RGB-only SwiftNet model was selected as baseline. {Runtimes in milliseconds ($ms$) are calculated by an RTX2070 GPU.}}
\vskip-1ex
\label{tab:event_fuse}
\centering
\begin{tabular}{ccccccrrl}
\toprule
\textbf{Network} & \textbf{RGB} & \textbf{Event} & \textbf{Fusion} & \textbf{Event Bins} & \textbf{Params(M)} & \textbf{Runtime(ms)} & \textbf{Cityscapes} & \textbf{DADA-seg} \\
\midrule \midrule
SwiftNet~\cite{orsic2019swiftnet}    & $\checkmark$  &   &  &         & 11.816 & 8.1~(\textpm 0.17) & 69.2   & 20.1   \\ \midrule
SwiftNet~\cite{orsic2019swiftnet}    & & $\checkmark$   & & $B=1$    & 11.810 & 7.9~(\textpm 0.15) & 35.6   & 2.3    \\
SwiftNet~\cite{orsic2019swiftnet}    & & $\checkmark$   & & $B=2$    & 11.813 & 8.0~(\textpm 0.32) & 36.0   & 19.7   \\
SwiftNet~\cite{orsic2019swiftnet}    & & $\checkmark$   & & $B=10$   & 11.838 & 8.6~(\textpm 0.07)  & \textbf{38.4}   & \textbf{21.7}~\gbf{+1.6}   \\
SwiftNet~\cite{orsic2019swiftnet}    & & $\checkmark$   & & $B=18$   & 11.863 & 9.5~(\textpm 0.33) & 36.6   & 19.8   \\
\midrule
EDCNet   & $\checkmark$ & $\checkmark$ & S2D - EAM & $B=1$  & 16.955 & 13.9~(\textpm 0.19) & 68.3   & 16.7   \\
EDCNet   & $\checkmark$ & $\checkmark$ & S2D - EAM & $B=2$  & 16.958 & 13.9~(\textpm 0.19) & \textbf{68.4}   & \textbf{23.0}~\gbf{+2.9}   \\
EDCNet   & $\checkmark$ & $\checkmark$ & S2D - EAM & $B=10$  & 16.983 & 14.6~(\textpm 0.21) & 66.5   & 18.8   \\
EDCNet   & $\checkmark$ & $\checkmark$ & S2D - EAM & $B=18$  & 17.009 & 15.4~(\textpm 0.22) & 67.1   & 10.4   \\ \midrule

EDCNet   & $\checkmark$ & $\checkmark$ & D2S - EGM & $B=1$   & 12.012 & 51.7~(\textpm 0.90)  & 69.0   & 24.5   \\
EDCNet   & $\checkmark$ & $\checkmark$ & D2S - EGM & $B=2$   & 12.012 & 56.3~(\textpm 2.40) & \textbf{69.4}   & \textbf{28.3}~\gbf{+8.2}   \\ 
EDCNet   & $\checkmark$ & $\checkmark$ & D2S - EGM & $B=10$   & 12.012 & 52.7~(\textpm 2.18) & 68.8 & 22.9 \\
EDCNet   & $\checkmark$ & $\checkmark$ & D2S - EGM & $B=18$   & 12.013 & 52.8~(\textpm 2.27) & 68.8 & 24.5 \\
\bottomrule
\end{tabular}
\vskip-3ex
\end{table*}

\subsection{Performance Gap}\label{exp:performance_gap}
To quantitatively evaluate the robustness of semantic segmentation algorithms, existing accuracy- and efficiency-oriented models are tested on the target dataset, as shown in \cref{tab:domain_gap} and visualized in~\cref{fig:miou_gap}. For a fair comparison, when applicable, the results and model weights are provided by the respective publications. It can be seen that, although both lightweight models~\cite{romera2017erfnet}\cite{orsic2019swiftnet}\cite{poudel2019fastscnn}\cite{Howard_2019MobileNetV3} and large models~\cite{zhao2017PSPNet}\cite{chen2018deeplabv3plus}\cite{yuan2019OCRNet}\cite{fu2019dual}\cite{yin2020dnl}\cite{zhang2020resnest}\cite{Kirillov2019PanopticFPN} gain high accuracy in the source domain~(Cityscapes), they heavily depend on the consistency between training and testing data, which are all normal scenes. It thus hinders their generalization ability and leads to a large performance degradation once taken to the abnormal scenes. The performance degradation of networks on this accident dataset are around $50\%-60\%$, of which the $63.6\%$ decline of DNL~\cite{yin2020dnl} is the most serious, from $79.3\%$ to $15.7\%$. {Vision transformer-based methods including SETR~\cite{setr} and SegFormer~\cite{segformer} designed specifically for semantic segmentation have the capacity to gather long-range context from early layers, but also suffer from huge performance drops in accident scenes.} One noticeable insight is that Fast-SCNN~\cite{poudel2019fastscnn} has the highest mIoU {among the CNN-based methods} on DADA-seg with $26.3\%$ despite only having a score of $69.1\%$ in Cityscapes. Overall, the large gap shows that semantic segmentation in accident scenarios is an extremely challenging task for these top-performance models.

\subsection{Experimental Settings}
For efficiency reasons, we choose ResNet-18~\cite{he2016resnet} as the backbone and the main architecture from SwiftNet~\cite{orsic2019swiftnet}, which is also selected as the baseline model in this work. All models in this subsection are constructed with the encoder-decoder structure and implemented on two GTX1080Ti GPUs with CUDA 11.0, CUDNN 8.0.4 and PyTorch 1.7.0. These models are trained with the Adam optimizer~\cite{kingma2014adam} with a learning rate initialized to $4{\times}10^{-4}$. The cosine annealing learning rate scheduling strategy~\cite{Loshchilov2017SGDRSG} is used to dynamically adjust the learning rate during the training process. The minimum learning rate of the last epoch is fixed as $1{\times}10^{-6}$. The weight decay of the learning rate is set to $1{\times}10^{-4}$. We use the weights of the ResNet-18 model pre-trained on ImageNet~\cite{russakovsky2015imagenet} to initialize the RGB encoder, and use the Kaiming initialization method~\cite{he2015kaiming_init} to initialize the whole Event branch as well as the decoder. For parameters of the pre-trained RGB encoder, we update them with a $4\times$ smaller learning rate than the initialized parameters of the event encoder and the decoder, and at the same time apply a $4\times$ smaller weight decay of the learning rate. The data augmentation operation includes a random scaling factor between $0.5$ and $2$, random horizontal flipping, and random cropping with an output resolution of $1024{\times}512$. In the end, we trained the model for $200$ epochs with a batch size of $2$ per GPU.

\subsection{Experiments on EDC}\label{exp:event_fusion}
\textbf{Quantitative Analysis.}
{To verify the benefits of extracting dynamic context from event-driven data, we perform the ablation study in EDCNet of two different modes (EAM and EGM) and four various event time bins B=\{$1$, $2$, $10$, $18$\}.} As shown in~\cref{tab:event_fuse}, we train the SwiftNet with RGB only from scratch as a baseline, which obtains $20.1\%$ mIoU in the target domain. {Starting exploring event data with event-only SwiftNet, where the event data are processed alone from sparse to dense without RGB images, the time bin settings larger than $1$ can achieve a similar performance as the baseline, and the setting of B=$10$ attains the mIoU of $38.4\%$ in the source domain and $21.7\%$ in the target domain.} This indicates that the event data has certain interpretability for the segmentation of driving scenes. {Although the accuracy of the Event-only model in the source domain is lower than the RGB-only model, the accuracy in the target domain is close or slightly better. According to such smaller domain gaps, the event data is more domain-consistent than RGB images. This is another insight leading us to develop the Event-aware Domain Adaptation method, in which the domain-consistent event data is processed as auxiliary information with RGB images to further bridge and align two domains. Benefiting from the EAM module fusing RGB images with event data, our EDCNet (EAM) obtains an mIoU improvement of $+2.9\%$ in the target domain based on the setting of B=$2$. At the same time, it maintains a high performance in the source domain.}

Regarding various time bin settings, one reason for the performance drop in the case of B=$1$ is that all events embedded in a single time bin will cause a problem of dragging behind images of moving objects and then negatively impact RGB features. However, in higher time bins, such as $10$ and $18$, the events generated in a short interval are dispersed to more bins, which then it leads to insufficient events in a single bin. Therefore, we verify the moderate event representation (B=$2$) as an effective time bin setting of our EDCNet to perform RGB-Event fusion, because others are either redundant or too sparse for the RGB image.

Likewise, we conduct the second group of experiments on EDCNet based on EGM in~\cref{tab:event_fuse}. Benefiting from the gating mechanism of the EGM module in the D2S extraction way, the EDCNet in $4$ different time bins performs more robustly than the baseline and generally achieves a considerable improvement. Especially, EDCNet in B=$2$ brings over an $+8.2\%$ gain in mIoU in the target domain when compared with the baseline, which is in line with the aforementioned time bin setting. This EDCNet obtains a $28.3\%$ mIoU score on the accident dataset and it outperforms more than $20$ representative accuracy- and efficiency-oriented {CNN models} listed in~\cref{tab:domain_gap}. We leave the Transformer methods in our future work of Transformer-based domain adaptation.

To summarize briefly, the input from the two data domains are obviously complementary as: \textit{(i)} When event cameras will not be triggered in static scenes, conventional cameras can perfectly capture the entire scene and provide sufficient textures. \textit{(ii)} When RGB cameras puzzle over adverse scenes, \textit{i.e.} fast-moving objects or low lighting environments, the event data can provide auxiliary information, which is particularly important for the segmentation of accident scenes. 

\begin{figure}[!t]
    \centering
    \includegraphics[width=0.99\columnwidth]{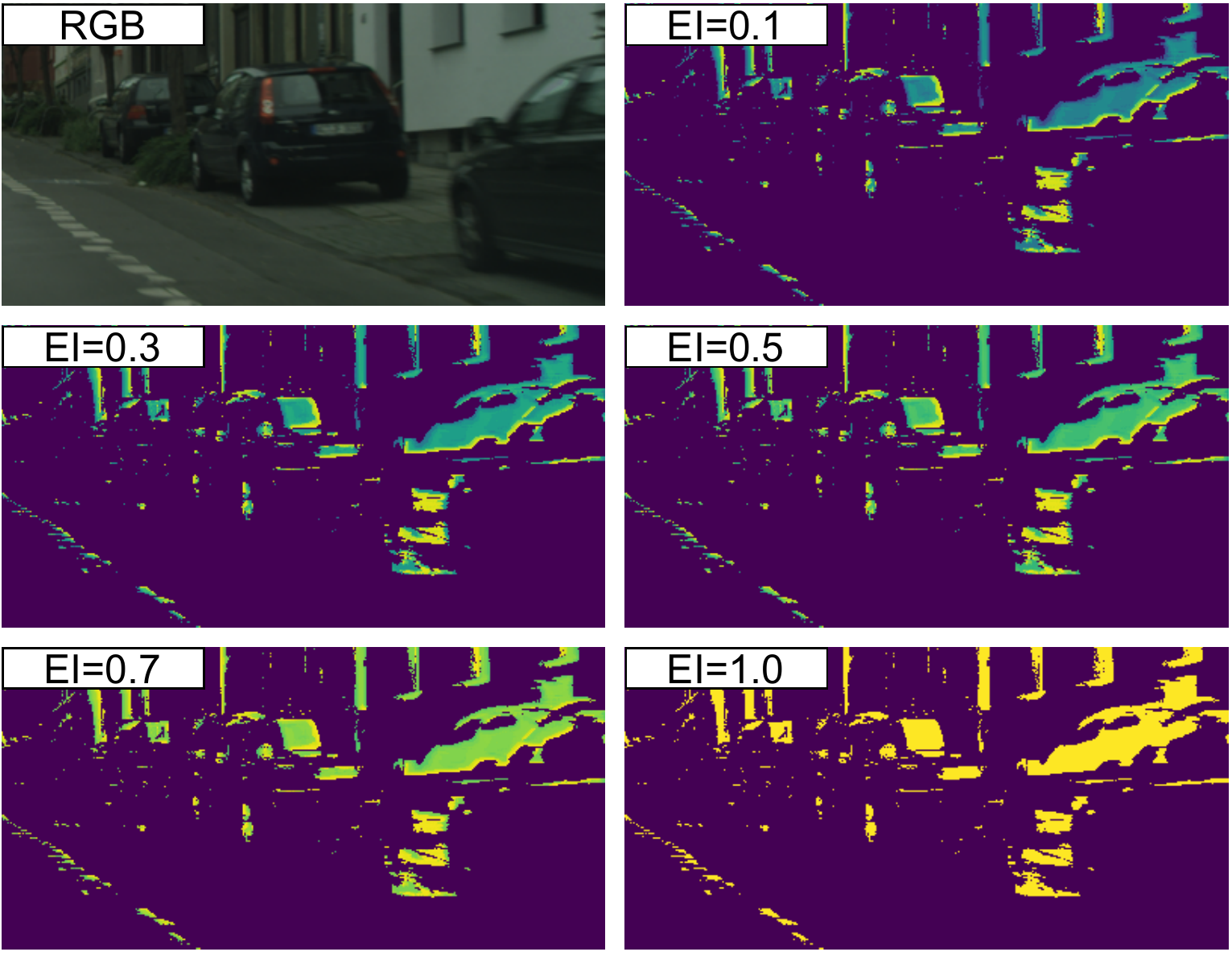}
    \vskip -1ex
    \caption{\small Visualization of various settings of event intensity~(EI).}
    \label{fig:vis_event_intensity}
\vskip -3ex
\end{figure}

\textbf{Ablation Study of Event Intensity.}
Since the intensity of the accumulated event in the resulting Image Warped Events~(IWE) affects the score of the reward functions and the performance of downstream event-based applications~\cite{stoffregen2019event}, we conduct an ablation study on event intensity~(EI) during the preprocessing of the synthesized event-based data. While the default setting of EI within a same time range in EventGAN~\cite{zhu2019eventgan} is set as $0.1$, we modify it to different settings with EI${\in}\{0.1, 0.3, 0.5, 0.7, 1.0\}$. The event-based data in EI=$1.0$ has the largest intensity. The visualizations of event-based data in various settings are displayed in Fig.~\ref{fig:vis_event_intensity}. Each model is based on the aforementioned D2S event fusion mode, and the event time bin is set as B=$2$. The models are trained on a single GPU with a batch size of $2$ and tested on the DADA-seg dataset. As the results shown in Table~\ref{tab:ablate_event_intensity}, EDCNet has a further improvement, elevating mIoU to $21.98\%$, when EI increases from $0.1$ to $0.3$. However, if the intensity is too high or reaches the maximum (EI=$1.0$), the performance will decrease. One reason is that the differentiation of event data in the temporal dimension is limited in a higher EI setting. For instance, when EI=$0.7$, most of the event pixels in the accumulated IWE will reach the maximum value (yellow pixels in Fig.~\ref{fig:vis_event_intensity}), which leads to excessive homogeneity, thereby losing dynamic  information of event data.

\begin{figure*}[t]
    \centering
    \includegraphics[width=0.99\textwidth]{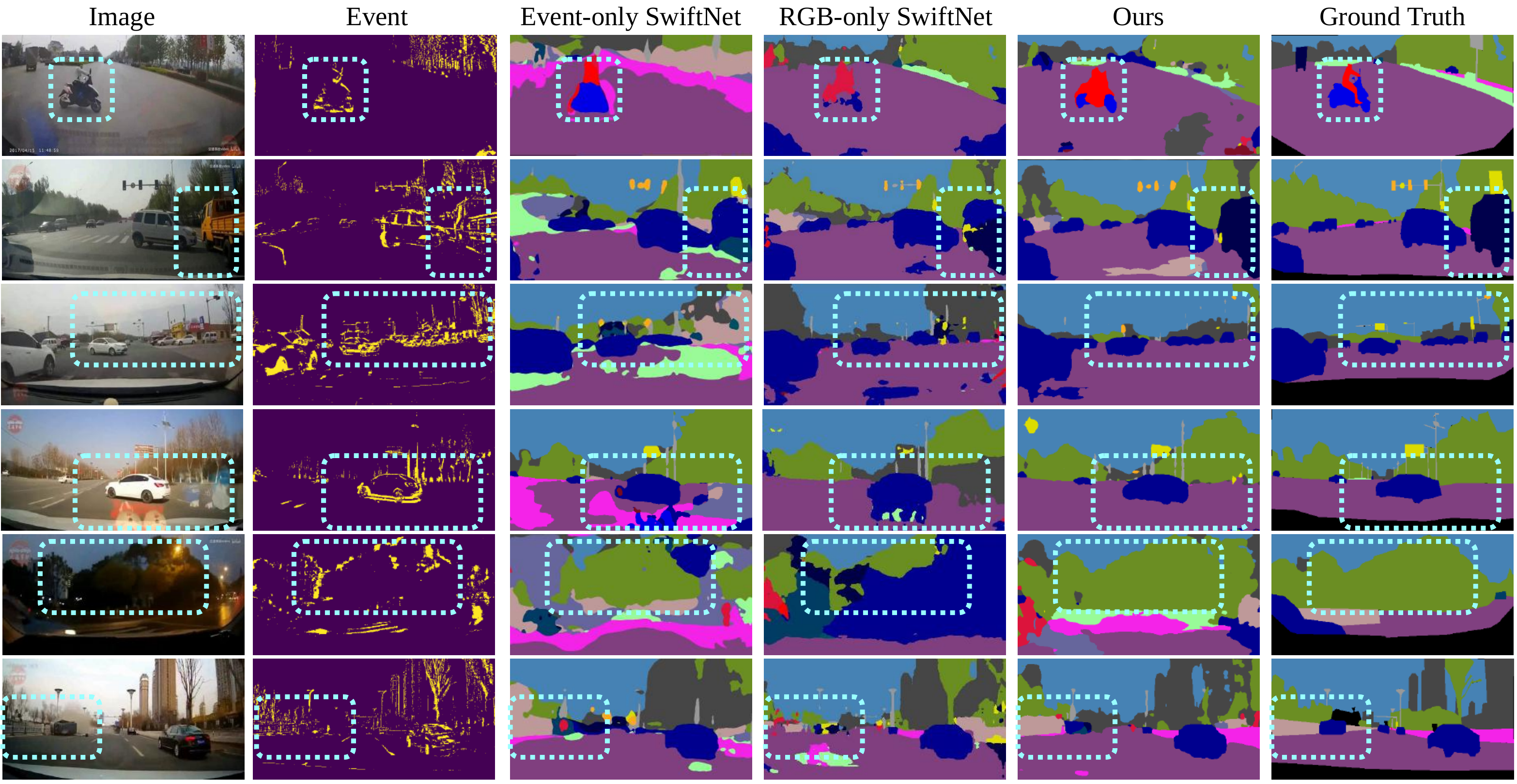}
    \vskip -1ex
    \caption{\small Contrastive examples between the Event-only SwiftNet, RGB-only SwiftNet and our EDCNet. From top to bottom are different accident scenarios: motorcyclist collision, truck and car dashing, windshield occlusion, car collision at night time, and an initial accident with an overturned car. The abnormalities are highlighted in light-blue boxes. Zoom in for a better view.}
    \label{fig:comparison_swiftnet}
\vskip -3ex
\end{figure*}

\textbf{Qualitative Analysis.}
{The comparison of accident scenes segmentation in \cref{fig:comparison_swiftnet} demonstrates that our EDCNet performs significantly better by fusing events and RGB images in those challenging situations. The region of interest in each row is hightlighted in light-blue dashed boxes. As shown in the first four rows of~\cref{fig:comparison_swiftnet}, our EDCNet concentrates on the motion information, especially the foreground objects, while the Event-only and RGB-only SwiftNet are both confused with the object categories, such as \emph{motorcycle} and \emph{motorcyclist} in the first row, \emph{car} and \emph{truck} in the second row, \textit{etc.} In the fifth row, our model improves the segmentation at the night-time scene. However, segmentation of night scenes is still challenging, although our method greatly benefits from event data, in contrast to the baseline. In the last row, a case of the initial accident scene is presented, where \emph{an overturned car} is lying on the road after collision with the fence, where our approach also clearly performs more robustly than the baseline and other compared methods.} 

\begin{table}[t]
\setlength{\tabcolsep}{17pt}
\caption{\small Ablation study of event intensity~(EI). Models are trained on Cityscapes and evaluated on DADA-seg dataset.}
\vskip-1ex
\label{tab:ablate_event_intensity}
\centering
\begin{tabular}{c|ccc}
\toprule
\textbf{Event Intensity} & \textbf{Acc} & \textbf{mIoU} & \textbf{fwIoU} \\
\midrule \midrule
0.1 & 41.06 & 20.80 & \textbf{49.04} \\
0.3 & \textbf{43.66} & \textbf{21.98} & 48.86 \\
0.5 & 41.00 & 21.16 & 48.08 \\
0.7 & 39.55 & 20.25 & 46.57 \\
1.0 & 40.98 & 20.33 & 45.89 \\
\bottomrule
\end{tabular}
\vskip -4ex
\end{table}

\begin{figure}[t]
    \centering
    \includegraphics[width=0.90\columnwidth]{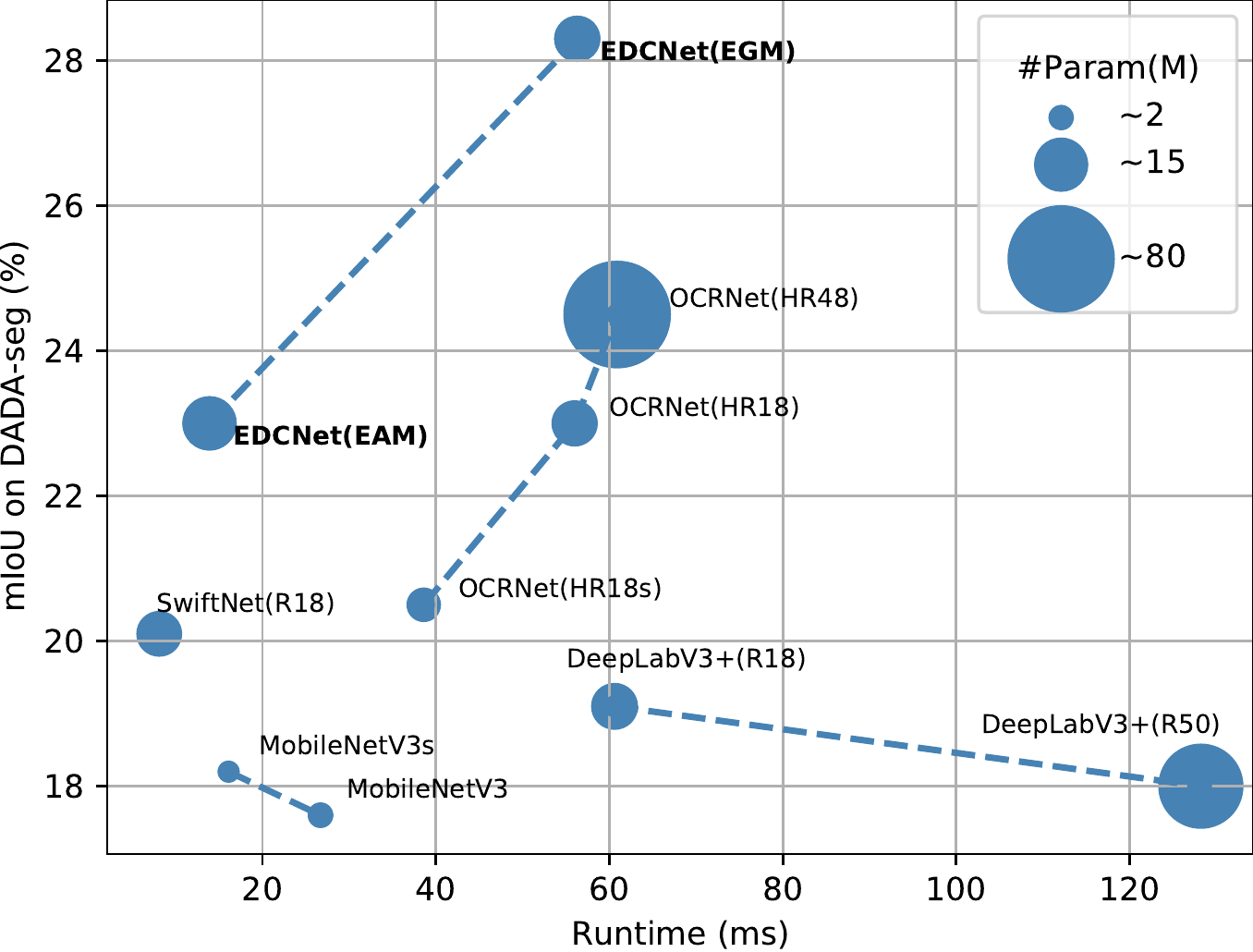} 
    \vskip -1ex
    \caption{\small {Illustration of trade-off between accuracy, runtime, and parameters across different models. Runtime~($ms$) and mIoU accuracy~($\%$) on DADA-seg dataset are depicted in x- and y-axis, while the size of the blue dot represents the number of parameter~($M$).}}
    \label{fig:runtime}
\vskip -4ex
\end{figure}

\subsection{Model Efficiency Analysis}

{To maintain the efficiency of model similar as SwiftNet~\cite{orsic2019swiftnet}, all of our EDC models are constructed based on the lightweight backbone ResNet-18~\cite{he2016resnet}. In order to compare the running time of all models uniformly and fairly, we performed calculations with $100$ cases based on the same setting on a workstation with an RTX2070 GPU, and finally listed the average runtime in milliseconds~($ms$) and the corresponding standard deviation~($\textnormal{\texttt{\textpm}} \Delta ms$). As shown in the Runtime column of \cref{tab:event_fuse}, although our EDC model in the S2D mode contains two completely symmetrical RGB and Event branches, it can still maintain a sufficient real-time speed and only increases the running time by $6ms$ in general, compared to the original SwiftNet which only has a single branch. However, the design of EDC model in the D2S mode aims to obtain more representative events from dense RGB images, and thus the feature maps in the Event branch are operated at the full-resolution scale. Compared with S2D, the runtime increases to around $50ms$, but the number of parameters (see \emph{Params} column in \cref{tab:event_fuse}) of D2S is much less than that of S2D, which is almost equal to the single-branch SwiftNet model.}

{In Fig.~\ref{fig:runtime}, we visualize a comparison between our EDCNet and existing representative semantic segmentation networks~\cite{chen2018deeplabv3plus}\cite{yuan2019OCRNet}\cite{orsic2019swiftnet}\cite{Howard_2019MobileNetV3}, regarding the trade-offs of runtime~($ms$), model accuracy~(mIoU), and parameters~($M$). First, our EDCNet in both modes (\textit{i.e.} EAM and EGM) has few parameters (around $15M$). EDCNet in EAM mode has similar runtime (around $20ms$) compared to the efficiency-oriented models, such as SwiftNet~\cite{orsic2019swiftnet} and MobileNet~\cite{Howard_2019MobileNetV3}, but its accuracy in DADA-seg dataset clearly outperforms theirs. Within the $60ms$ runtime range, EDCNet in EGM mode achieves much higher performance with a similar or a smaller number of parameters, compared with DeepLabV3+(R18)~\cite{chen2018deeplabv3plus} and OCRNet(HR48)~\cite{yuan2019OCRNet}.}

\subsection{Benchmarking and Comparison with the State-of-the-Art}
\label{sec:sota}

\begin{table}[t]
\setlength{\tabcolsep}{14pt}
\renewcommand{\arraystretch}{0.8}
\caption{\small Comparison of the EDCNet and the SwiftNet on diverse datasets. The SwiftNet is trained only with RGB images as baseline, while our EDCNet is set as the D2S-EGM mode. The Merge3 dataset is combined from Cityscapes, KITTI-360, and BDD3K.}
\vskip -1ex
\label{tab:exp_dataset}
\centering
\begin{tabular}{@{}c|ccl@{}}
\toprule
\textbf{Dataset} & \textbf{Network} & \textbf{Source} & \textbf{DADA-seg} \\
\midrule \midrule
\multirow{2}{*}{Cityscapes~\cite{cordts2016cityscapes}} & SwiftNet~\cite{orsic2019swiftnet}    & 69.2  & 20.1 \\
& EDCNet   & 69.4 & 28.3~\gbf{+8.3}  \\ \midrule
\multirow{2}{*}{BDD3K~\cite{yu2020bdd100k}} & SwiftNet~\cite{orsic2019swiftnet}    & 30.6  & 23.9 \\
& EDCNet   & 36.5 & 28.6~\gbf{+4.7}  \\ \midrule
\multirow{2}{*}{KITTI-360~\cite{xie2016semantic}} & SwiftNet~\cite{orsic2019swiftnet}    & 45.2  & 13.7 \\
& EDCNet   & 46.6 & 16.1~\gbf{+2.4}  \\ \midrule
\multirow{2}{*}{ApolloScape~\cite{wang2019apolloscape}} & SwiftNet~\cite{orsic2019swiftnet}    & 61.8  & 16.7 \\
& EDCNet   & 58.8 & 19.5~\gbf{+2.8}  \\ \midrule
\multirow{2}{*}{Merge3} & SwiftNet~\cite{orsic2019swiftnet}    & 50.3  & 28.5 \\
& EDCNet   & 61.4& 32.4~\gbf{+3.9}  \\
\bottomrule
\end{tabular}
\vskip -4ex
\end{table}

\begin{table*}[ht]
\centering
\small \setlength{\tabcolsep}{2.5pt}
\caption{\small Performance comparison of domain adaptation strategies, where \textit{f} and \textit{i} represent the feature and image level transfer between the source and target domain. 
The results of \textit{Source} and \textit{Target} are tested with $1024{\times}512$ resolution, while \textit{Target$\dag$} and \textit{Foreground classes} are with $512{\times}256$. To clearly showcase the effect of the event-aware branch, the per-class IoU(\%) of ten foreground classes of \textit{Target} result are listed: \textit{Traffic Light, Traffic Sign, Pedestrian, Rider, Car, Truck, Bus, Train, Motorcycle}, and \textit{Bicycle}. Note that the target dataset does not have any \textit{Train}. Our EDCNet-Event model is adapted on feature and image levels by fusing event data in D2S mode with the EGM module, while the EDCNet-DOF is by fusing dense optical flow data.}
\label{tab:comparison_model}
\begin{tabular}{@{}ll|rrrrrrrrrr|ccc|ccc|ccc@{}}
\toprule
\multirow{2}{*}{\textbf{Network}} & \multirow{2}{*}{\textbf{Level}}  & \multicolumn{10}{c|}{\textbf{Foreground classes}} & \multicolumn{3}{c|}{\textbf{Target}$\dag$} & \multicolumn{3}{c|}{\textbf{Source}} & \multicolumn{3}{c}{\textbf{Target}} \\ \cmidrule{3-21}
& & TLi & TSi & Ped & Rid & Car & Tru & Bus & Tra & Mot & Bic & Acc & mIoU & fwIoU  & Acc & mIoU & fwIoU & Acc & mIoU & fwIoU   \\ \midrule \midrule
CLAN~\cite{luo2019CLAN}  & -   & 15.2 & 5.3 & 4.0 & 3.4 & 32.6 & 8.8 & 28.8 & - & 4.2 & 0.1 & 34.0 & 19.4 & 45.5 & 56.3 & 43.7 & 77.2 & 28.1 & 16.8 & 38.3 \\
CLAN~\cite{luo2019CLAN}  & f   & 17.2 & \textbf{21.5} & 8.4 & 6.3 & 63.5 & 33.4 & 33.1 & - & 3.7 & 6.2 & 46.3 & 31.7 & 67.2 & 70.4 & 62.4 & 87.0 & 40.1 & 28.8 & 63.8 \\
CLAN~\cite{luo2019CLAN}  & f+i & 17.0 & 20.0 & 9.4 & 5.2 & 64.3 & 36.8 & 35.9 & - & 5.6 & \textbf{7.7} & 47.3 & 32.4 & 66.3 & 73.2 & \textbf{64.8} & 87.3 & 39.4 & 28.2 & 60.6 \\
\midrule
EDCNet-DOF & f+i & \textbf{18.1} & 17.7 & 9.5 & 8.1 & 64.3 & 34.8 & 34.9 & - & 5.1 & 7.3 & \textbf{48.3} & \textbf{33.4} & \textbf{69.6} & 71.6 & 62.9 & 87.4 & 40.9 & 29.2 & 64.3 \\
EDCNet-Event & f+i & 17.0 & 19.5 & \textbf{10.0} & \textbf{8.8} & \textbf{65.6} & \textbf{39.5} & \textbf{39.7} & - & \textbf{6.1} & 7.0 & 48.2 & 33.1 & 68.2 & \textbf{73.2} & 63.9 & \textbf{87.5} & \textbf{42.1}  & \textbf{30.0}  & \textbf{64.5}  \\
\bottomrule
\end{tabular}
\vskip -3ex
\end{table*}

\textbf{On Diverse Datasets.}
For extensive verification of the event-driven dynamic context, we conduct comparisons of our EDCNet with the RGB-only SwiftNet on other datasets, such as ApolloScape~\cite{wang2019apolloscape}, KITTI-360~\cite{xie2016semantic}, and BDD3K~\cite{yu2020bdd100k}. For the event data synthesis on all these datasets, we perform the aforementioned selection process. The contrastive results are shown in~\cref{tab:exp_dataset}. The \textit{Merge3} dataset is merged from the Cityscapes, KITTI-360, and BDD3K datasets, because these three datasets have the same labels with all $19$ classes. In general, our EDCNet is capable of improving the segmentation robustness by extracting and integrating dynamic context from the event-based data. Due to the adverse driving scenes included in the BDD3K dataset, both SwiftNet and EDCNet have a smaller performance gap between the source and target dataset. {Based on the data diversity of BDD3K}, our EDCNet gains a $+5.9\%$ improvement on the source domain and a $+4.7\%$ improvement on the DADA-seg dataset, compared to SwiftNet. On the KITTI-360 and ApolloScape datasets, our model has also considerable improvements in the target domain, precisely $+2.4\%$ and $+2.8\%$, respectively. In particular, the mIoU of our EDCNet on the Merge3 dataset reaches the highest score $32.4\%$. {The resulting EDCNet clearly gets ahead all the CNN methods and  successfully exceeds state-of-the-art transformer-based models.} Overall, the results show that our proposal is consistently and significantly effective for enhancing the reliability of accident scene segmentation. Besides, verifying on diverse datasets demonstrates the reproducibility of our proposed methods if one wants to deploy with or extend to other cameras or datasets. On the other hand, the DADA-seg dataset is sufficiently diverse (see Table~\ref{tab:condition_distribute}) to cover the evaluation of models trained from different source domains.

\subsection{Experiments on EDA}\label{Subsection4_2}
As mentioned before, apart from the event-aware domain adaptation, we aim to boost the robustness of semantic segmentation by performing different Unsupervised Domain Adaptation (UDA) strategies on the large-scale unlabeled data of our DADA-seg dataset. Additionally, this ablation study includes the comparison between fusing the event-based data and the dense optical flow. To compare diverse strategies, based on the recent model CLAN~\cite{luo2019CLAN}, our EDCNet method is performed on two different levels, \textit{i.e.} feature and/or image level. For an extensive quantitative analysis, we have adopted three different metrics~\cite{long2015FCN}, namely pixel accuracy (Acc), mean intersection over union (mIoU), and frequency weighted intersection over union (fwIoU), as shown in \cref{tab:comparison_model}.

\textbf{Quantitative Analysis.}
Firstly, the CLAN model was adapted from the virtual to the real domain, \textit{i.e.}, from the GTA5~\cite{richter2016gta5} to the Cityscapes dataset. This pre-trained model is tested directly on our DADA-seg dataset without any adjustments, also named source-only CLAN. It gains the mIoU of $16.8\%$ with a $1024{\times}512$ resolution and $19.4\%$ with a $512{\times}256$ resolution, respectively. Note that here a smaller resolution input results in a higher accuracy in the target domain. There are two main reasons: \textit{(i)} images of DADA-seg are originally with a low resolution, and \textit{(ii)} a smaller resolution can obtain a larger receptive field with wider context understanding, which indicates that correct classification is more critical in accident scenes than delineating the boundaries. Afterwards, we train the CLAN model from scratch by adapting from Cityscapes to DADA-seg to verify the feature-level and feature-image-level domain adaptation, whose are termed as $f$ and $f+i$ shortly in \textit{Level} column of~\cref{tab:comparison_model}, whereas the latter obtained a higher mIoU of $64.8\%$ in the source domain. Finally, based on the CLAN model design, our EDCNet in the EDA strategy obtains the highest performance on all three metrics in the DADA-seg dataset, and achieves the top accuracy of $30.0\%$ in mIoU, $42.1\%$ in Acc, and $64.5\%$ in fwIoU at the resolution with $1024{\times}512$. In order to understand the impact of event fusion, we list the per-class IoU results of all $10$ foreground classes in \cref{tab:comparison_model}. This demonstrates that the foreground classes indeed benefit more from event data, which is consistent with our assumptions.

\textbf{Comparison with Optical Flow.}
In order to compare the dynamic context from other data modalities, we replace the event-based data with Dense Optical Flow~(DOF) simulated by the Farneback function~\cite{farneback2003DOF}. For a fair comparison based on the same sparsity of data, we only utilize the traditional method to generate optical flow data. Under the same architecture and setting of our EDCNet, the EDCNet-DOF model also obtains accuracy improvements. Nevertheless, our EDCNet-Event model performs better in the foreground classes. Although both data are synthesized, motion features with higher time resolution can be extracted from event-based data to boost the foreground segmentation. Besides, compared to optical flow, event cameras actively capture the intensity change and also have a high dynamic range to enhance perception in low-light conditions, which better conforms with our EDCNet design for improving driving safety by fusing event-driven dynamic context. 

\begin{table*}[!t]
\renewcommand\arraystretch{1.2}
\footnotesize
\setlength{\tabcolsep}{4pt}
\begin{center}
\caption{Comparison with state-of-the-art domain adaptation methods. Results of mIoU and per-class IoU are calculated in the resolution of $1024{\times}512$ on DADA-seg dataset.}
\vskip-1ex
\label{tab:compair_sota_da}
\begin{adjustbox}{max width=\textwidth}
\begin{tabular}{ l | c | rrrrrrrrrrrrrrrrrrr}
\toprule[1pt]
\textbf{Method} & \rotatebox{90}{mIoU} &  \rotatebox{90}{Road} &  \rotatebox{90}{Sidewalk} &  \rotatebox{90}{Building} & \rotatebox{90}{Wall} &  \rotatebox{90}{Fence} &  \rotatebox{90}{Pole} & \rotatebox{90}{Traffic Light} &  \rotatebox{90}{Traffic Sign}&  \rotatebox{90}{Vegetation} &  \rotatebox{90}{Terrain} &  \rotatebox{90}{Sky} & \rotatebox{90}{Person} &  \rotatebox{90}{Rider} & \rotatebox{90}{Car} &  \rotatebox{90}{Truck}& \rotatebox{90}{Bus}& \rotatebox{90}{Train}& \rotatebox{90}{Motorcycle}&  \rotatebox{90}{Bicycle}\\
\hline
\hline
CLAN~\cite{luo2019CLAN} & 28.76 & 79.80 & 18.61 & 51.56 & 8.32 & 13.60 & 15.51 & 17.15 & 21.51 & 63.20 & 21.99 & 80.53 & 8.37 & 6.32 & 63.47 & 33.43 & 33.12 & - & 3.69 & 6.21 \\
BDL~\cite{li2019bidirectional} & 29.66 & 81.44 & 19.18 & 57.18 & 8.61 & 16.26 & 14.65 & 8.78 & 16.77 & 66.60 & 26.83 & 85.87 & 10.51 & 7.16 & 65.45 & 35.18 & 34.78 & - & 2.71 & 5.57 \\
FDA~\cite{yang2020fda} & 24.45 & 67.83 & 15.36 & 39.99 & 4.28 & 12.98 & 16.08 & 6.93 & 18.26 & 57.01 & 9.71 & 64.50 & 10.30 & 9.49 & 59.39 & 16.84 & 39.35 & - & 11.28 & 4.94 \\
SIM~\cite{wang2020differential} & 26.85 & 79.13 & 16.93 & 56.79 & 4.69 & 12.38 & 16.50 & 7.47 & 14.04 & 65.78 & 18.17 & 87.27 & 11.04 & 3.83 & 61.33 & 22.99 & 28.21 & - & 0.52 & 2.96  \\
EDCNet (ours) & \textbf{29.97} & 80.23 & 19.51 & 52.02 & 6.43 & 14.68 & 16.19 & 17.03 & 19.50 & 65.39 & 21.69 & 79.84 & 9.95 & 8.82 & 65.60 & 39.51 & 39.73 & - & 6.09 & 7.03 \\
\bottomrule[1pt]
\end{tabular}
\end{adjustbox}
\end{center}
\end{table*}

\begin{figure}[t]
    \centering
    \includegraphics[width=0.99\columnwidth]{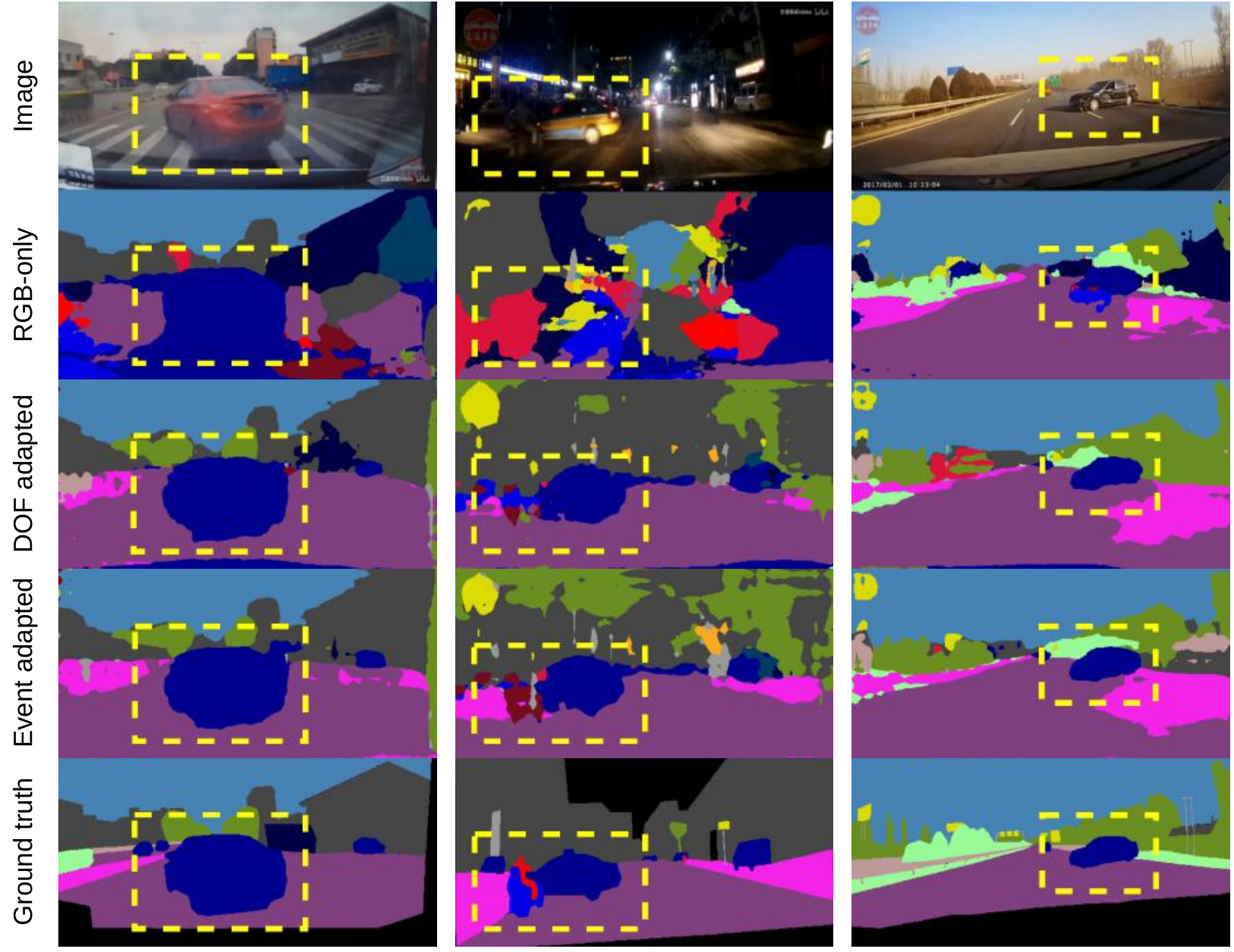}
    \caption{ \small Comparison of segmentation results between the RGB-only baseline, EDCNet-DOF, and EDCNet-Event on DADA-seg dataset. These accident scenes are more complicated by involving motion blur, night-time condition, and initial accidents during driving, respectively. {The fast moving or adverse foregrounds are highlighted by yellow dashed boxes. Zoom in for a better view.}}
    \label{fig:results_comparison}
\vskip -3ex
\end{figure}

\begin{figure*}[t]
    \centering
    \includegraphics[width=0.99\textwidth]{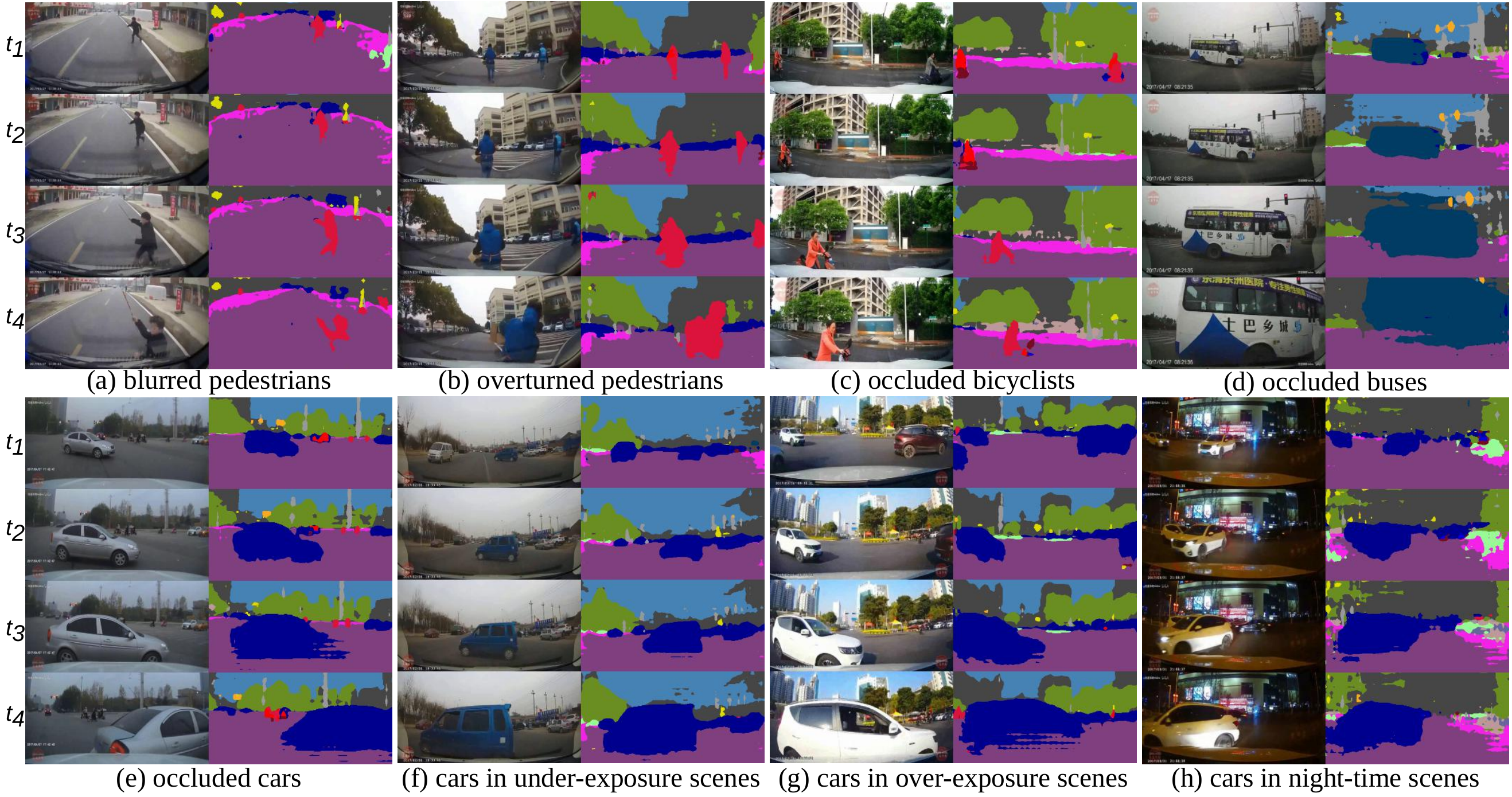}
    \caption{ \small More semantic segmentation results with the respective sequence during a traffic accident of the DADA-seg dataset. The columns correspond to the input images and output predictions of our EDCNet. {The objects involved in the collision from (a) to (e) include pedestrians, bicyclists, cars, and buses. Accident scenes from (f) to (h) include under-exposure, over-exposure and night-time conditions.}}
    \label{fig:results_CLAN}
\vskip -2ex
\end{figure*}

\textbf{Comparison with State-of-the-art DA Methods.}
{Table~\ref{tab:compair_sota_da} shows the comparison between our EDCNet and some state-of-the-art domain adaptation methods~\cite{wang2020differential}\cite{luo2019CLAN}\cite{li2019bidirectional}\cite{yang2020fda}. Results including mIoU and IoU for each category are tested on the target domain, \textit{i.e.} DADA-seg dataset. For a fair comparison, these models are trained based on the default settings recommended by the referenced works, to perform domain adaptation from Cityscapes to DADA-seg. Compared with the baseline model CLAN, our model achieves a gain of $1.21\%$ by fusing event data, which reveals our aforementioned insight, the domain-consistency of event-based data. The EDCNet obtains respective $0.29\%$, $5.52\%$, and $3.12\%$ gain in mIoU, compared with BDL~\cite{li2019bidirectional}, FDA~\cite{yang2020fda}, and SIM~\cite{wang2020differential}. Compared to the baseline, in addition to the mIoU improvement, it shows much higher accuracy in foreground objects (such as \emph{car}, \emph{truck}, \emph{bus}, and \emph{bicycle}). This result is consistent with our assumption that event data are beneficial for foreground segmentation in domain adaptation.}

\textbf{Qualitative Analysis.}
A qualitative comparison of semantic segmentation results between the RGB-only SwiftNet, EDCNet-DOF, and EDCNet-Event is shown in~\cref{fig:results_comparison}. {These three samples are selected based on complicated driving accident scenes, where images from left to right cover the motion-blur condition, the night-time, and the initial accident ahead. The adverse foreground objects are highlighted via yellow dashed boxes in~\cref{fig:results_comparison}. In the motion-blur scenario, when the RGB-only baseline is confused between the foregrounds and backgrounds, our EDCNet accurately segments the foreground object (\textit{i.e.} \emph{the blurred car}). Apart from blurred scenes, our EDCNet obtains better performance in the night scene where the RGB-only SwiftNet almost fails. Likewise, an incident car lying horizontally in an initial accident ahead is precisely segmented by our EDCNet.
Additionally, \cref{fig:results_CLAN} shows more segmentation results of our EDCNet-Event model on DADA-seg in a sequence manner. From sequence (a) to sequence (e), our model can handle the motion blur, overturns, object occlusions caused by \emph{car-pedestrian, car-bus, or car-car collisions}. The segmentation sequences of three accident cases in different lighting conditions (under-exposure, over-exposure, and night-time) reveal that our model can perform semantic segmentation robustly against various adverse situations.} 
These indicate that our EDCNet model significantly stabilizes the segmentation in normal and abnormal scenes by fusing event data, especially for the blurred foreground objects during traffic accidents.

\begin{figure*}[t]
    \centering
    \includegraphics[width=0.99\textwidth]{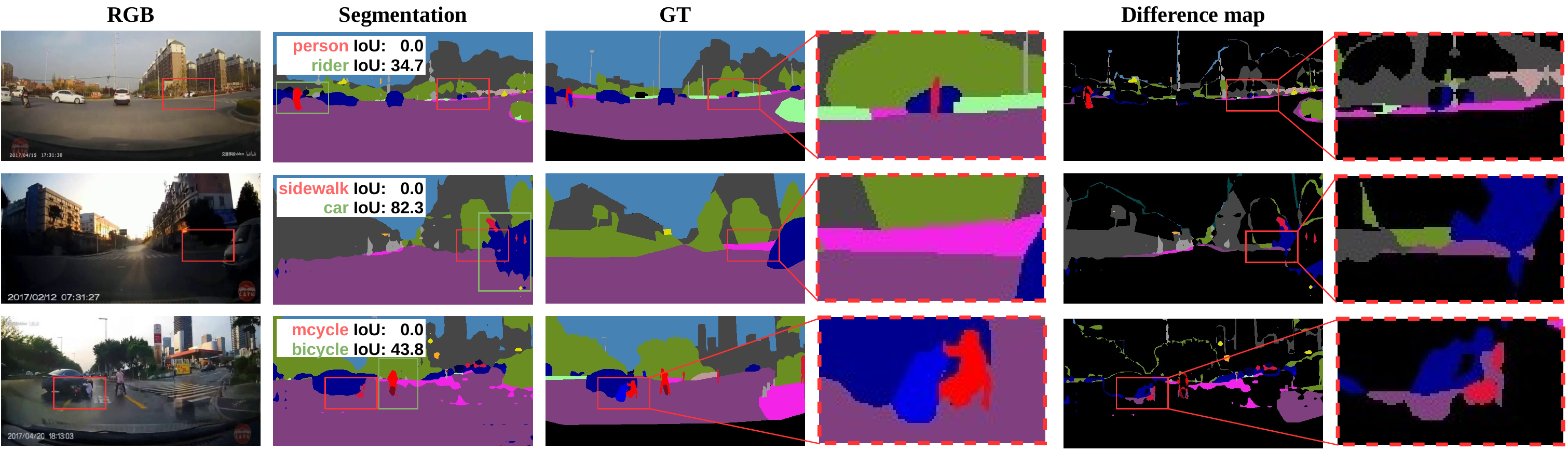}
    \vskip -1ex
    \caption{\small Failure analysis of segmentation results. From left to right are the input image, the segmentation result of EDCNet, the ground truth (with zoom-in region), and the difference map (with zoom-in region). The difference map is generated by subtracting the true positive pixels from the segmentation result \textit{w.r.t.}} the ground truth. ``mcycle'' is the motorcycle class for short. Zoom in for a better view.
    \label{fig:failure_case}
\vskip -3ex
\end{figure*}

\textbf{Failure Analysis.}
In order to comprehensively analyze the performance of our EDCNet on different cases, some failure segmentation examples are visualized in Fig.~\ref{fig:failure_case}. In the first row, the segmentation of the pedestrian at a long distance failed, but the model can still segment the rider at a mid-range distance with an IoU of $34.7\%$. In the second failed case, the static sidewalk as the background cannot be segmented in the low-light environment, but the model can still obtain the dynamic context of the moving car via fusing the event data. In the last row, as the motorcycle collided with the car and was overturned, the model was confused and failed in segmenting this motorcycle, but the bicycle next to it can be segmented with an IoU of $43.8\%$. It is worth noting that segmenting traffic accident scenes in the real world is still tough. To obtain a better perception of highly dynamic driving scenes, panoramic cameras can be used to obtain a wider field of view (\textit{i.e.}, $360^\circ$) to perceive pedestrians or cars dashing from the side of ego-vehicle in advance. In addition, fusing event data in transformer-based models that are capable of capturing long-distance context is a potential solution.

\section{Conclusions}
In this work, we present the accident semantic segmentation task, aiming to enhance the road safety by perceiving traffic accident scenes during driving.
To this end, we build its relevant evaluation dataset DADA-seg with pixel-wise annotations, which serves as a benchmark to assess the robustness and applicability of semantic segmentation algorithms for IV systems.
As an initial solution to segment those accidental scenarios, we mainly explore event-driven dynamic context from the generated event data and investigate the possibility of applying event cameras in future vehicles. As a new sensing module, the event camera can be fused with other sensors, such as a normal camera discussed in this work. Imported as a novel perception module in the general pipeline of autonomous vehicles, our proposed semantic segmentation methods mainly focus on improving the perception of the abnormality happened in a traffic accident, so as to provide better scene understanding information for subsequent path planning and control modules.
Based on the characteristic of event data, we construct the multimodal segmentation model \emph{EDCNet} by fusing the RGB with event-based data through different modes and attention mechanisms. Our extensive experiments on various datasets show that dynamic context captured from event data is capable of complementing the RGB image features with temporal information. We further boost the performance by merging diverse datasets and using the EDA strategy. Thus, under extreme driving situations, such as in scenes with motion blur and low illuminations, event cameras are helpful to better perceive the environment and to improve image analysis. Even though our experiments are somewhat limited by the use of synthetic events due to the lack of corresponding event data in common annotated datasets, we have observed consistent and large accuracy gains.

Eventually, the analysis of traffic accident scenarios is still scarce, especially the dense environment understanding \textit{i.e.} semantic segmentation. On the one hand, it is still lacking a large-scale dataset with pixel-wise annotation. On the other hand, the traffic accident containing various adverse factors and challenging situations all belong to edge cases of driving scenarios. Thus, our current segmentation performance on those accident scene still has large development space. For this purpose, the large-scale unlabeled data in the DADA-seg dataset may be explored through other learning paradigms, such as contrastive learning.
According to the failure analysis, $360^\circ$ cameras and fusing event data in transformer-based model would be alternative techniques to gain robust segmentation, which we leave in our future work.
Moreover, an equally intriguing possibility is the accident prediction based on the combination of video semantic segmentation and event regression algorithms, which is a potential approach to avoid traffic hazards and further ensure road traffic safety.

\bibliographystyle{IEEEtran}
\bibliography{main}

\end{document}